\documentclass[authoryear]{elsarticle}
\pdfoutput=1
\usepackage[bookmarks=false]{hyperref}

\usepackage{scrextend}
\usepackage{fullpage}
\usepackage{amssymb}
\usepackage{amsmath}
\usepackage{mathtools}
\usepackage{bm}
\usepackage{tabularx} 
\usepackage[linesnumbered,ruled]{algorithm2e}
\usepackage{csvsimple, booktabs, siunitx, xstring}
\usepackage{import}
\usepackage[skip=10pt]{caption}
\usepackage{subcaption}
\usepackage{mathtools}
\usepackage{makecell}
\usepackage{array, multirow}
\usepackage{arydshln}
\usepackage{float}
\usepackage{tikz}

\journal{ArXiv}

\newcommand{\fatline}{\Xhline{4\arrayrulewidth}}

\SetCommentSty{mycommfont}
\graphicspath{{pics/}}

\newcommand{\eg}{\textit{e.g. }}
\newcommand{\ie}{\textit{i.e. }}  
\newcommand{\comment}[1]{}
 
\newcommand{\KSP}[0]{{\bf KSPTrack}}
\newcommand{\KSPOPT}[0]{{\bf KSPTrack$^{opt}$ }}

\newcommand{\KSPnb}[0]{{KSPTrack }}
\newcommand{\KSPOPTnb}[0]{{KSPTrack$^{opt}$ }}

\newcommand{\EEL}[0]{{\bf EEL}}
\newcommand{\PSVM}[0]{{\bf P-SVM}}
\newcommand{\GS}[0]{{\bf Gaze2Segment}}
\newcommand{\DL}[0]{{\bf DL-prior}}

\newcommand{\EELnb}[0]{{EEL}}
\newcommand{\PSVMnb}[0]{{P-SVM}}
\newcommand{\GSnb}[0]{{Gaze2Segment}}
\newcommand{\DLnb}[0]{{DL-prior}}

\newcommand{\blue}[1]{#1}

\usepackage{numcompress}
\biboptions{authoryear}

\begin{document} 
\begin{frontmatter}

\title{Iterative multi-path tracking for video and volume \\ segmentation with sparse point supervision}

\author[]{Laurent Lejeune\corref{mycorrespondingauthor}}
\cortext[mycorrespondingauthor]{Corresponding author}
\ead{laurent.lejeune@artorg.unibe.ch}

\author{Jan Grossrieder}
\author{Raphael Sznitman}

\address{
Ophthalmic Technology Laboratory,
ARTORG Center, University of Bern,
Murtenstrasse 50, 3008 Bern,
Switzerland}

\begin{abstract}
\label{sec:orgheadline1}
Recent machine learning strategies for segmentation tasks have shown great ability when trained on large pixel-wise annotated image datasets. It remains a major challenge however to aggregate such datasets, as the time and monetary cost associated with collecting extensive annotations is extremely high. This is particularly the case for generating precise pixel-wise annotations in video and volumetric image data. To this end, this work presents a novel framework to produce pixel-wise segmentations using minimal supervision. Our method relies on 2D point supervision, whereby a single 2D location within an object of interest is provided on each image of the data. Our method then estimates the object appearance in a semi-supervised fashion by learning object-image-specific features and by using these in a semi-supervised learning framework. Our object model is then used in a graph-based optimization problem that takes into account all provided locations and the image data in order to infer the complete pixel-wise segmentation. In practice, we solve this optimally as a tracking problem using a K-shortest path approach. Both the object model and segmentation are then refined iteratively to further improve the final segmentation. We show that by collecting 2D locations using a gaze tracker, our approach can provide state-of-the-art segmentations on a range of objects and image modalities (video and 3D volumes), and that these can then be used to train supervised machine learning classifiers. 
\end{abstract}


\end{frontmatter}

\newcommand\mybold[2]{%
    \IfEqCase{#2}{
      {True}{\textbf{#1}}
      {TRUE}{\textbf{#1}}
      {False}{#1}
      {FALSE}{#1}
      }
}

\newcommand\printType[2]{\ifthenelse{\equal{#1}{KSP}}{\multirow{3}{*}{#2}}{}}

\newcommand{\convmethod}[1]{%
    \IfEqCase{#1}{%
        {vilar}{\PSVMnb}%
        {gaze2}{\GSnb}%
        {mic17}{\EELnb}%
        {wtp}{\DLnb}%
        {KSP}{\KSPnb}%
        {KSPopt}{\KSPOPTnb}%
    }[\PackageError{convmethod}{Undefined option to convmethod: #1}{}]%
}%

\newcommand\dataall[2]{\csvreader[no head,
  respect backslash=true,
  respect rightbrace=true,
  respect leftbrace=true,
  late after line=\ifthenelse{\equal{\smethods}{vilar}}{\\\cdashline{2-10}}{\\},
  late after last line=\\\hline,
  before first line=,
  column count=24,
  ]
  {#1}{
    1=\smethods,
    2=\fa,
    3=\pra,
    4=\rca,
    5=\bolda,
    6=\fb,
    7=\prb,
    8=\rcb,
    9=\boldb,
    10=\fc,
    11=\prc,
    12=\rcc,
    13=\boldc,
    14=\fd,
    15=\prd,
    16=\rcd,
    17=\boldd,
    18=\boldall,
    19=\fmean,
    20=\fstd,
    21=\prmean,
    22=\prstd,
    23=\rcmean,
    24=\rcstd}
  {\printType{\smethods}{#2} & \convmethod{\smethods} & $\mybold{\fa}{\bolda}$ & $\mybold{\fb}{\boldb}$ & $\mybold{\fc}{\boldb}$ & $\mybold{\fd}{\boldd}$ & ~~ & $\mybold{\fmean}{\boldall} \pm \fstd$ & $\prmean \pm \prstd$& $\rcmean \pm \rcstd$}
    }

\newcommand{\convloss}[1]{%
    \IfEqCase{#1}{%
        {L2}{\text{U-Net}}
        {overfeat}{\text{OverFeat}}
        {L2prior}{\text{Image-object}}
    }[\PackageError{convloss}{Undefined option to convloss: #1}{}]%
}%
\newcommand\printTypeLoss[2]{\ifthenelse{\equal{#2}{L2prior}}{\multirow{3}{*}{#1}}{}}
\newcommand\losscomp{
  \csvreader[no head,
    tabular=cccccccc,
    column count=17,
  late after line=\ifthenelse{\equal{\themethod}{L2prior}}{\\\cdashline{1-8}}{\\},
    late after last line=\\\hline,
    table head=
      \toprule \multirow{2}{*}{Type} & \multirow{2}{*}{Features} & \multicolumn{4}{c}{F1} & ~~ & F1 \\
       &  & 1 & 2 & 3 & 4 & ~~ & mean $\pm$ std \\ \toprule,
    filter expr={
          test{\ifnumgreater{\thecsvinputline}{1}}}
    ]
  {tables/ksp_losses.csv}{1=\thetype,
    2=\themethod,
    3=\bolda,
    4=\meana,
    5=\stda,
    6=\boldb,
    7=\meanb,
    8=\stdb,
    9=\boldc,
    10=\meanc,
    11=\stdc,
    12=\boldd,
    13=\meand,
    14=\stdd,
    15=\boldall,
    16=\meanall,
    17=\stdall}
    {\printTypeLoss{\thetype}{\themethod} & \convloss{\themethod} & $\mybold{\meana}{\bolda} \pm \stda$ & $\mybold{\meanb}{\boldb} \pm \stdb$ & $\mybold{\meanc}{\boldc} \pm \stdc$ & $\mybold{\meand}{\boldd} \pm \stdd$& ~~ & $\mybold{\meanall}{\boldall} \pm \stdall$}
    }

    \newcommand\dataallfeatextr[1]{
  \csvreader[no head,
    tabular=cccccc,
    column count=16,
    table head=
              \toprule Types & ScP & VGG & U-Net & U-Net / prior\\\toprule,
    late after last line=\\\hline,
    filter expr={
          test{\ifnumgreater{\thecsvinputline}{1}}}
    ]
  {#1}{1=\smethods,
    2=\mscp,
    3=\sscp,
    4=\escp,
    5=\mvgg,
    6=\svgg,
    7=\evgg,
    8=\munet,
    9=\sunet,
    10=\eunet,
    11=\munetg,
    12=\sunetg,
    13=\eunetg}
    {\smethods & \mscp $\pm \sscp \cdot 10^{-\escp}$ & \mvgg $\pm \svgg \cdot 10^{-\evgg}$ & \munet $\pm \sunet \cdot 10^{-\eunet}$ &  \munetg $\pm \sunetg \cdot 10^{-\eunetg}$}
    }

\section{Introduction}
\label{sec:intro}

At its core, semantic segmentation is tasked with associating pixels, or voxels, of an image with a label that corresponds to a meaningful category. As a fundamental problem in medical image computing, an impressive amount of research on the topic has been conducted in recent years, spanning methods that segment tumors in MRI volumes~\citep{zikic2014,menze15}, airways from chest CT scans~\citep{miyawaki17}, vessels in retinal scans~\citep{pilch12} or mitochondria in electron microscopes~\citep{seyed13} to name a few.

To this day, the vast majority of state-of-the-art segmentation approaches rely heavily on supervised machine learning frameworks~\citep{sweeney14,menze15}, and in particular deep learning~\citep{garcia17}, to produce excellent segmentation results. In general, these methods depend on large amounts of training examples to learn complex prediction models. Critically, most of these models are trained using pixel-wise annotations associated with training examples. While highly effective, the cost of acquiring such pixel-wise annotations for training machine learning methods is often overlooked and yet a central limiting factor for aggregating extremely large training datasets. This is particularly the case for video and 3D image data where annotations are extremely costly (\ie days per video sequence). This in turn negatively impacts the capacity to train high-performing segmentation models, as the number of training samples remains relatively small. 

To reduce the burden of producing pixel-wise annotations, a number of semi-supervised concepts have been proposed such as active learning~\citep{KonSznFua15}, domain adaption~\citep{tzeng17} and crowd-sourcing~\citep{mavandadi12}. Alternatively, a number of recent methods propose to infer pixel-wise segmentation directly from image labels (\ie the image contains a tumor) by leveraging strong object or shape priors~\citep{menze10}. In effect these methods attempt to refine segmentations from pre-trained neural networks for generic object classes~\citep{su15} or trained attention models~\citep{kingma14}. While extremely promising, such methods still only produce coarse segmentations and remain ill suited for training complex prediction models. 

At the same time, the method by which annotations are provided has important practical implications in terms of convenience for the annotator and can also greatly speed-up the annotation process~\citep{ferreira12}. For example, providing tumor pixel segmentations in volumetric data would be infeasible if each pixel were to be specified individually. Instead, there is an important body of work that has considered alternative user-interaction mechanisms. For instance,~\citep{KonSznFua15} used 3D image planes to specify the boundary between image backgrounds and objects in 3D modalities. Scribbles of positive and negative image regions were also shown to be effective in speeding up annotation generation~\citep{roberts11}. Even more efficient, was the use of 2D points to sparsely annotate image data~\citep{bromiley14,bearman16}. Such 2D point supervision is interesting as it can be produced by manually clicking on images from 3D volumes or video sequences, or by using a gaze tracker to passively record 2D coordinates of the object as they are viewed~\citep{yunpeng13,khosravan16,lejeune17}. This latter strategy is promising as it holds the potential to annotate at high framerate, but is challenging due to limited, if any, information regarding the background. To overcome this, previous methods that generate segmentations from 2D supervision have relied on strong assumptions on the object size, the background scene, or clouds of 2D locations, whereby limiting usability.

\begin{figure}[t!]
\centering
\includegraphics[width=0.99\textwidth]{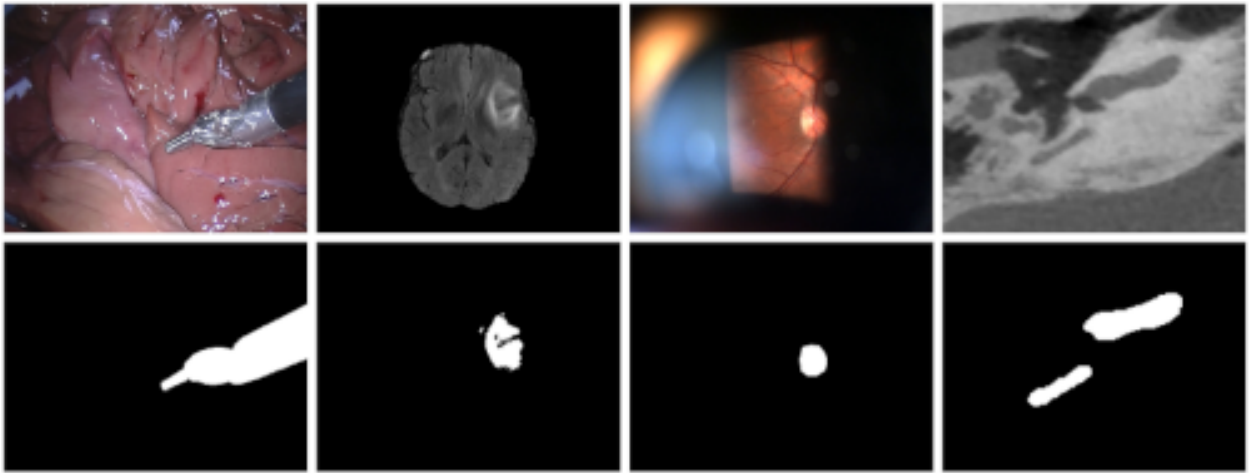}
\caption{Example of objects of interest in different video and volumetric modalities. The bottom row shows the pixel wise segmentation for each case: From left to right: A surgical instrument during an endoscopic procedure, a 3D MRI scan containing a brain tumor, an optic disc seen from a slit lamp microscope and a cochlea cross-section in a 3D CT scan.}
\label{fig:fig1}
\end{figure}

To generalize the use of sparse 2D point supervision to infer segmentations in a given image volume or sequence, we propose a novel framework that avoids the need to assume much about the object and only requires 2D object locations to be specified. In particular, we assume that the object of interest is compact and always present in the image, but can have arbitrary size regardless of the image modality considered. With the goal of segmenting this object of interest, our method first builds an object appearance model from the provided 2D locations. We then construct a 3D graph over the entire image data, from which the complete object segmentation is estimated in each frame. To do this, the graph is optimized in a multi-source, single sink max-flow setting that uses the object model and we show that this problem can be solved optimally using a K-shortest path approach. We then iterate this optimization procedure using the previously obtained result to update our object model and thus refine the produced segmentations. In order to achieve consistent segmentations from such sparse annotations regardless of the image type or object, we also propose to use image and object specific features that are learned from the image data. This is achieved by using a Convolutional Neural Network (CNN) and a novel loss function that takes into account the 2D image locations. Combined, we show that our framework provides a significant improvement over existing methods on a number of varied datasets (see Fig.~\ref{fig:fig1}). We show that our results are not only more similar to traditional hand segmentations to those produced by state-of-the-art methods, but that they only induce mild reductions in performance when used to train prediction models. Beyond this, we show that by using a low-cost gaze tracker to generate supervised 2D locations, a user can generate annotations at high framerate.

In the following sections, we begin by describing existing methods closest to our work. We then introduce our framework in Sec.~\ref{sec:overview} and describe its detail thereafter. In Sec.~\ref{sec:results}, we show how our method performs on a number of tasks and datasets, as well as compared to existing methods.


\section{Related Works}
\label{sec:related_works}
While semi-supervised methods for segmentation tasks encompass a wide range of applications and settings~\citep{Chapelle2006}, we briefly discuss a number of methods that are related to the present paper. In particular, we focus on graph-based and learning-based methods, as well as methods that leverage different user-input mechanisms. 

\noindent
{\bf Graph-based methods:} 
Graph-based methods are well studied in both the computer vision and medical imaging communities. The seminal work of~\cite{Boykov2006} first introduced an efficient and optimal method for binary segmentation using both object and background appearence models. GrabCut and other variants~\citep{rother04,yu2014} further improved the approach using iterative optimizations. At their core, these methods rely on object and background models, computed from provided supervision, to segment objects. More recently, the approach of~\cite{karthikeyan13} extracts visual {\it tracklets} by combining gaze inputs from multiple individuals and optimizes a patchwork of locations using a Hungarian algorithm to globally extract bounding boxes that are then refined using GrabCut. In particular, by leveraging crowds of users to provide pointwise indications of object of interest, the method effectively produces segmentations from clouds of points. In contrast, in the approach we propose, only a single point per frame within the object of interest is given. This is similar to the work of~\cite{khosravan16}, who make use of saliency maps~\citep{koch98} derived from gaze locations in CT scans to segment lung lesions. The saliency maps serve as object and background models (assuming bounds on the lesion sizes) in a graph cut optimizer.

\noindent
{\bf Semi-supervised learning methods:} A wide range of semi-supervised learning methods are related to our present work. Given that in our setting, only positive examples consisting of parts of the object are provided, our problem is closely related to transductive learning~\citep{Burges13,Guyon17} and more specifically P(ositive)-U(nlabeled) learning~\citep{Li2005,Kiryo2017}. In such cases,  only part of the positive set is labeled in addition to a large amount of unlabeled data. To tackle this setting, most methods focus on providing more adapted loss functions during training or leveraging priors to constrain the ensuing classifier. 

Early on and also using a gaze tracker,~\cite{vilarino07} suggested a P-U learning setting to detect polyps from endoscopic video frames. This approach bares some semblance to ours, except that we explicitly take into account temporal information by means of a graph to further constrain our segmentation. At the same time, unlike their approach, we do not assume that the object is of a given size. Along this line, \cite{lejeune17} considered a P-U setting by explicitly learning a classifier using a loss function that takes into account the uncertainty associated with unlabeled samples. These uncertainties are derived from gaze locations while Probability Propagation~\citep{zhou04} is used to estimate unknown samples. Within a deep learning framework,~\cite{bearman16} suggested learning a CNN using gaze information as well as a strong object prior in order to improve convergence of their network. The method performs well on natural images of complex scenes, as the objectness prior is learned from a large corpus of natural images. Similarly, FusionSeg~\citep{jain17} used a deep learning approach with an initial object outline to segment object boundaries in video sequences. This approach, which is highly related to tracking, combines both motion and appearance to track the object with limited user interaction.

\noindent
{\bf User-input models:} Given the wide use of machine learning, the extensive research on user-input methods and interactive algorithms, is by no means surprising. Beyond traditional polygon outlining, scribbling has been proposed to annotate faster. 2D point locations, either on individual images or in video streams has also been shown to be effective when providing coarse information in extremely fast amounts of time~\citep{Papadopoulos17}. 

Related to the work here, gaze trackers have received an increasing amount of attention given that the technology has greatly improved over the last decade and seen a strong reduction in cost~\citep{soliman16,mettes16,bearman16}. In these works, gaze information provides a form of sparse annotations to train machine learning classifiers extremely quickly. In particular, large amounts of annotations can be accumulated by crowds of individuals observing natural video data for example. In the context of medical imaging, gaze locations have also been investigated to see how image annotation could be performed~\citep{sadegh09}, or how pathologies could be identified by a limited number of viewings of video or volumetric image data~\citep{vilarino07,khosravan16}. Unfortunately, most approaches so far have only been shown to work in extremely limited scenarios (\eg one type of object in a single modality). In our work, we show how object segmentations can be computed by using a gaze tracker to collect 2D locations of the object at framerate, in a single pass, without collecting or assuming information on the background scene, the object size or its motion speed. This allows our approach to be highly generic and effective on a variety of image modalities and object types. 


\section{Overview and problem formulation}
\label{sec:overview}

We now present our method that takes as input an image sequence (or an image volume) containing a single object of interest and produces a pixel-wise segmentation of this object for all frames. In general, we assume that at most one object of interest is in the sequence and that part of the object is visible in each frame. For each frame in the sequence, we assume that a 2D location (\ie a pixel) within the object is provided by a user.
Hence, while we are interested in determining the entire object segmentation, the provided information only specifies local and compact regions of the object. These provided locations may be spatially disjoint and can refer to different or the same areas of the object. For this reason, our approach treats the task as a tracking problem where the image regions specified by the 2D locations must be jointly and coherently tracked so to recover the complete object segmentation. To do this, our approach hinges on two components. 

The first is a strategy to characterize the object of interest by using the provided image sequence and associated 2D locations. This is achieved by learning a classifier in a transductive fashion. In particular, we resort to bagging a set of decision trees. Instead of combining the aforementioned classifier with hand-crafted or learned features from large datasets, we learn features explicitly from the considered image sequence while using the 2D locations as a soft prior. This is achieved by training a U-Net architecture~\citep{ronneberger2015} as an autoencoder in combination with a loss function that takes into account known object locations. By using the image features from this network, the classifier can then be used to assess the likelihood of image regions belonging to the object of interest. 

The second component considers each specified 2D location as a potential target to track. To segment the object from these locations, we construct a graph over all compact image regions (\ie superpixels) in the image sequence.
The object segmentation is then inferred with a network flow optimization strategy whereby each of the provided 2D locations correspond to flow sources and we use the object likelihood to establish a series of costs between adjacent edges in our graph. We show that this graph can be optimized exactly and efficiently using a K-shortest path approach. To further improve the segmentation, we update our classifier using the previously found segmentation and repeat the K-shortest path optimization to produce an improved segmentation. This process is iterated until convergence of the final produced segmentation.

\comment{ 
\begin{figure}[!ht]
  \centering
\begin{tikzpicture}
  \centering
    \node[anchor=south west,inner sep=0] (image) at (0,0) {\includegraphics[width=0.7\textwidth]{pics/pipeline}};
    \begin{scope}[x={(image.south east)},y={(image.north west)}]
        \node at (0.083,0.760)
    {Sec.~\ref{sec:features}};
        \node at (0.61,0.760)
    {Sec.~\ref{sec:optimization}};
    \end{scope}
\end{tikzpicture}
\caption{Full pipeline of our method.
  (Top) The dataset consists of a video/volumetric sequence and a set of 2D locations. (Middle) Our approach is divided in two main blocks: (Green frame) A feature learning phase where each pixels are assigned a feature vector. After a superpixel segmentation step, we assign to each superpixel a feature vector. (Red frame) A foreground model gives objectness likelihood to each superpixel. A MAP optimization problem then associates tracklets. Several iterations are performed and allow to re-train foreground and transition models. (Bottom) At the output, we obtain for each frame a binary segmentation of our object of interest.}
\label{fig:pipeline}
\end{figure}
}

By combining these two components, we show in our experiments that effective segmentations can be generated from extremely few 2D locations, without further prior assumptions on the image modality or the object of interest. 

We now briefly describe some notation that will be used throughout this paper and which are summarized in Table.~\ref{tab:notation}. Let the image sequence considered be denoted $\mathcal{I} = \{I_0,\ldots,I_T\}$ and let $\bm{g} = \{g_t\}_{t=0}^T$ with $g_t\in\mathbb{R}^2$ be a 2D pixel location in $I_t$. 
While we are ideally interested in a pixel-wise segmentation, we decompose each image as a set of superpixels so to reduce computational complexity. Given the volumetric nature of the problem considered, we opt to use 3D superpixels~\citep{chang13} to group similar pixels over multiple frames. We thus let $I_t$ be described by the set of $N_t$ non-overlapping superpixels $S_t=\{s^n_t\}_{n=0}^{N_t}$ and define the set of all superpixels across all images as $\mathcal{S}=\{S_t\}_{t=0}^T$. In addition, we assign to each superpixel $s_t^n$ an appearance feature vector $a_t^n$ and define $\bm{a}=\{a_t^n | t=0,\ldots,T\quad n=0,\ldots,N_t\}$. We denote the set $\mathcal{S}^p = \{s^n_t | g_t \in s^n_t, t=0,\ldots,T,n=0,\ldots N_t \}$ as all superpixels observed and the rest as $\mathcal{S}^u = \mathcal{S} \setminus \mathcal{S}^p$.
\begin{table}[t!]
\begin{center}
\begin{tabular}{l l l}
\toprule
Symbol & Description\\
\toprule
$T$  & Number of frames\\
$I_t$  & Image at time $t$\\
$g_t$  & Coordinates of 2D location at time $t$\\
$N_t$  & Number of superpixels at time $t$\\
$s_{t}^n$  & Superpixel $n$ at time $t$ \\
$a_{t}^n$  & Feature vector of $s_t^n$\\
$u_{t}^n$  & Histogram of oriented optical flow of $s_t^n$\\
$Y_{t}^n$  & Binary random variable that models objectness of $s_t^n$\\
$\rho_{t}^n$  & Probability of $s_t^n$ being object given the object model\\
$\mathcal{T}_{t}^n$  & Tracklet starting at time $t$ and superpixel $n$\\
$r_{t}^n$  & Centroid of $s_t^n$\\
$e_{t}^n$, $f_{t}^n$, $C_{t}^n$ & Edge, flow and cost for passing through $\mathcal{T}_t^n$\\
$e_t^{n,m}$, $f_t^{n,m}$, $C_t^{n,m}$  & Edge, flow and cost for linking $\mathcal{T}_t^n$ and $\mathcal{T}_{t+1}^m$\\
$e_t^{\mathcal{E},n}$, $f_t^{\mathcal{E},n}$, $C_t^{\mathcal{E},n}$  & Edge, flow, and cost for entering the network from $\mathcal{T}_t^n$\\
$\tau_{\rho}$  & Threshold applied on edges $e_t^n$ according to $\rho_t^n$\\
$\tau_{u}$  & Threshold applied on edges $e_t^{n,m}$ according to $u_t^n$\\
$\tau_{trans}$  & Threshold on edges $e_t^{i,j}$ and $e_t^{g,n}$ according to $u_t^n$\\
$R$  & Radius around 2D location (entrance), and tracklet transitions \\
$Z_{t}$  & Objectness prior at time $t$\\
$\sigma_{g}$  & Standard-deviation of objectness prior for feature extraction\\
\bottomrule
\end{tabular}
\end{center}
\caption{Notation summary}
\label{tab:notation}
\end{table}


\section{Transductive foreground model}
To build a model of the object appearance, we take a transductive learning approach. Here we follow a P-U learning scheme in which only a few positive samples are given along with a larger set of unknown samples. In practice for an image, we expect one superpixel to be annotated compared to hundreds of unobserved ones. While using Neural Networks would be effective for supervised binary segmentation problems, it is unclear what loss function one should minimize in a P-U regime. Instead, we propose to train a simple bagging classifier with novel features that are both image and object specific, and allow coarse superpixel regions to be characterized. 

\subsection{Probabilistic estimation by bagging}
\label{sec:foreground_model}
To build a prediction model, we train $M$ binary decision trees by using different data subsets. Each tree takes as input the feature vector $a_t^n \in \mathbb{R}^D$ characterizing the superpixel $s_t^n$ and estimates $Y_t^n \in \{0,1\}$, where $Y_t^n = 1$ if it belongs to the object and $0$ otherwise. For each tree, the entire positive set $\mathcal{S}^p$ is used for training, in addition to $|\mathcal{S}^p|$ randomly selected samples with replacement from the unlabeled set $\mathcal{S}^u$. The latter are treated as negative samples. The trees are then trained using the Gini impurity loss function~\citep{menze09}, with $\sqrt{D}$ randomly selected features considered at each node of a tree. Then for a given superpixel $s_t^n$, the probability that it is part of the object, $\rho_{t}^n = P(Y_t^n = 1 | a_{t}^n )$ can be computed by averaging the predictions over all $M$ trees.

\subsection{Image-object specific features}\label{sec:features}
In general, there are many different features that could be used in the above classifier. In this context however, we wish to use features that are effective for segmenting a specific object in a given image modality. That is, we are interested in learning features that are both image-object specific (IOS), but do not need to generalize to other unseen data. 

\begin{figure}[t]
\centering
\includegraphics[width=1\textwidth]{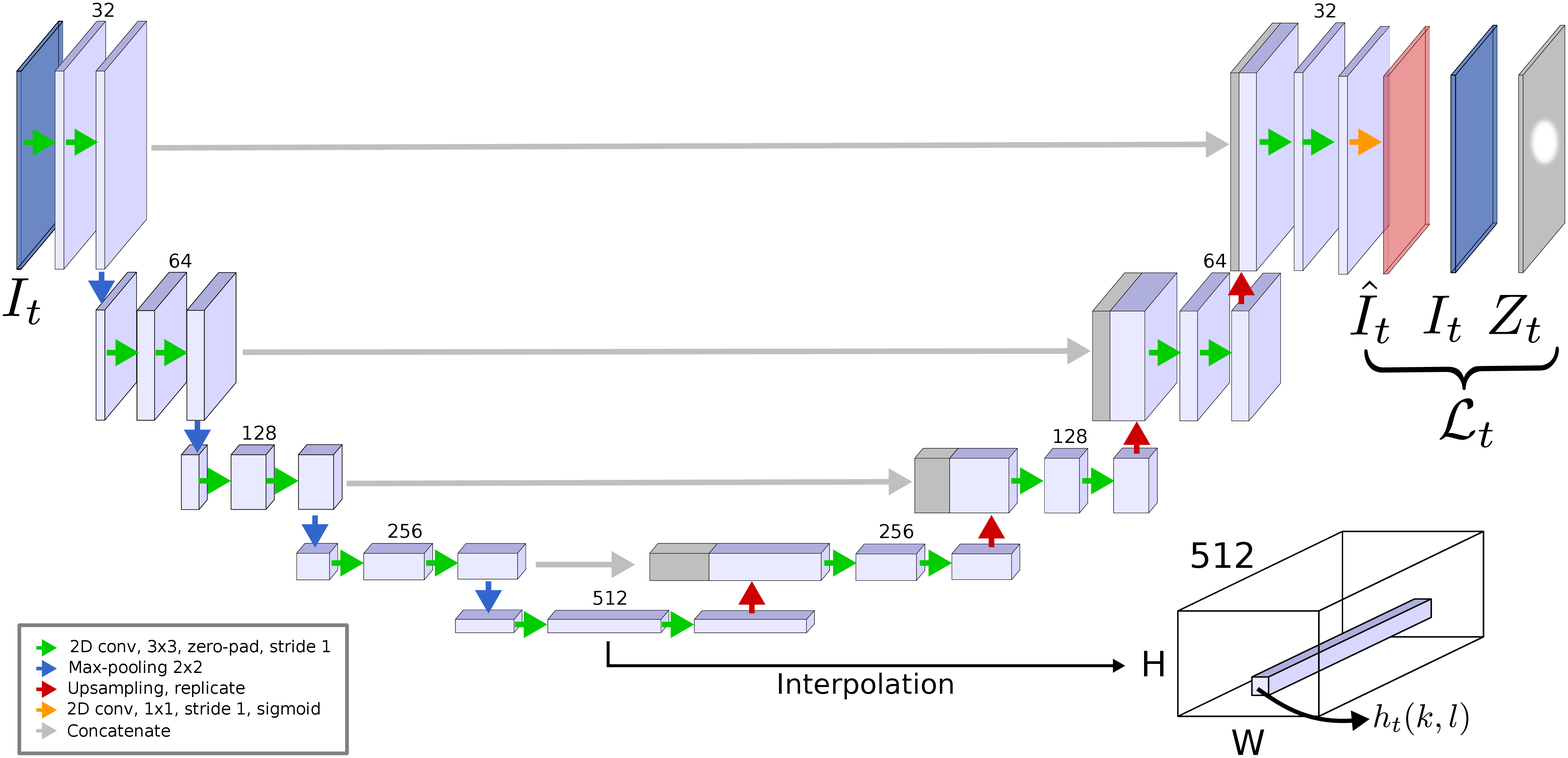}
\caption{Image-object specific features. The network is tasked to reconstruct the input image $I_t$ (dark blue). By means of a loss function $\mathcal{L}_t$, the reconstructed image, $\hat{I}_t$ (red), is strongly penalized at 2D locations provided by means of the soft prior, $Z_t$. At test time, the features $h_t(k,l)$ are extracted by interpolating the bottom layer to the original input size.
}
\label{fig:unet}
\end{figure}

To this end, we learn features by making use of an autoencoder neural network. As illustrated in Fig.~\ref{fig:unet}, we let the network take as input an image from a sequence and is tasked to predict the same image as output. In our network, we use three stacked convolutional layers with $3 \times 3$ filter with strides of $1$ per level in an encoding and decoding path. We also perform batch normalization with ReLU activations after each convolutional layer. The last layer is a convolutional layer with filter size $1 \times 1$ and a sigmoid activation. In practice, as our network has $4$ levels, where we first downscale the images to the nearest width and height divisible by $16$.

While such an autoencoder can be trained by minimizing the $L^2$-norm~\citep{vincent08}, we are interested in forcing the network to have strong performances on regions of the object of interest rather than on potentially irrelevant parts of the image. As such, we propose to impose an ``objectness prior", which we add to our loss function in the form of a soft constraint. Specifically, we define ${Z}_t \sim \mathcal{N}(g_t, \sigma_g^2\bm{1})$ to be a 2D weighted map of the same size as $I_t$. The mean of this image is centered on the location $g_t$ and has symmetric variance $\sigma_g^2$. We then integrate this map in our loss as
\begin{equation}
\mathcal{L} = \sum_{{I}_t \in \mathcal{I}} \sum_{k,l} {Z}_t(k,l) \| I_t(k,l) - {\hat{I}}_t(k,l)\|^2,
\label{eq:loss_features}
\end{equation}
\noindent
where $\hat{I}_t$ is the output of the network, $k$ and $l$ are pixel indices in a $W \cdot H$ sized image $I_t$. In effect, this loss penalizes incorrect reconstructions  more heavily on the regions that are known to be part of the object.

After training, a forward pass is performed on an image and features of dimension $512$ are extracted at the output of the deepest layer. These features correspond to a downscaled version of the input image which are then upscaled to the original image size using bicubic interpolation. The feature vector associated to a superpixel $s_t^n$ is then taken to be the mean over all pixels contained within it,
\begin{equation}
\begin{split}
  a_t^n = \frac{1}{|s_t^n|} \sum_{(k,l) \in s_t^n}h_t(k,l),
\end{split}
\label{eq:app_vector}
\end{equation}
\noindent
where $h_t(k,l) \in \mathbb{R}^{512}$ is the feature vector extracted at pixel $(k,l)$. As we will show in our experiments, this strategy generally improves the overall performance of our method.

\section{Segmentation by tracking} \label{sec:optimization}
Given the above local object model, we wish to provide a global strategy to infer an accurate segmentation of the object across all frames. As we make no assumption on the object of interest (\eg shape, color, motion, etc.), we \blue{hypothesise} that by tracking these local regions over the entire data volume, that a complete segmentation of the object can be coherently inferred. That is, we consider each region specified by a provided 2D location to be an individual target, that could potentially depict different parts of the same object. In what follows, we show how these different regions can be tracked optimally so as to provide the complete object segmentation.

\subsection{MAP Formulation}\label{sec:MAP}
To track the 2D locations as function of the object, we define $\bm{Y} = \{Y_t^n|\forall(t,n)\}$ as the set of all $Y$ labels.
As defined in Sec.~\ref{sec:overview}, $\bm{g}, \bm{a}$ are the grouping variables of the provided 2D locations and the extracted superpixel features, respectively.
We then define our segmentation problem as a Maximum a posteriori (MAP) optimization,
    \begin{equation}
    \begin{split}
    y^* = \arg \max_{y \in \mathcal{Y}} P(\bm{Y}=\bm{y}|\bm{a}, \bm{g}),
    \end{split}
    \label{eq:map}
    \end{equation}
\noindent
where $y^*$ is the sought out binary labels for all frames. Assuming that $Y_t^n$ is conditionally independent given the observed variables $\bm{a}$, we rewrite Eq.~\eqref{eq:map} as,
    \begin{equation}
    \begin{split}
    y^* = \arg \max_{y \in \mathcal{Y}} \prod_{\substack{m,n,t}} P(Y_t^{n}|\bm{a},\bm{g}) P(Y_t^{n}|a_{t-1}^{m})P(Y_t^{n}|a_t,g_t).
    \end{split}
    \label{eq:map2}
    \end{equation}
\noindent
In particular, the three terms of the decomposition of Eq.~\eqref{eq:map2} correspond to different aspects of the object appearance models. Concretely,
\begin{itemize}
\item[-]  {$P(Y_t^{n}|\bm{a},\bm{g})$} is modeled using our classifier (Sec.~\ref{sec:foreground_model}) and behaves as an object appearance model. 
\item[-] {$P(Y_t^{n}|a_{t-1}^{m})$} models the similarity between two superpixels in successive frames, so to describe how frame-to-frame probabilities propagate. 
\item[-] {$P(Y_t^{n}|a_t,g_t)$} models the likelihood that a given superpixel $s_t^n$ is visually similar to the one selected by the 2D location $g_t$. In practice, in the case where the object of interest is large and visually homogeneous, this term allows to initiate several tracks for a single given 2D annotation.
\end{itemize}

While optimizing Eq.~\eqref{eq:map2} appears complex, we show in the following section that it can be performed efficiently by means of an integer program formulation and a K-Shortest Path optimization. 

\subsection{Flow network formulation}
\label{sec:solving}
\begin{figure}[t]
\centering
\includegraphics[width=1\textwidth]{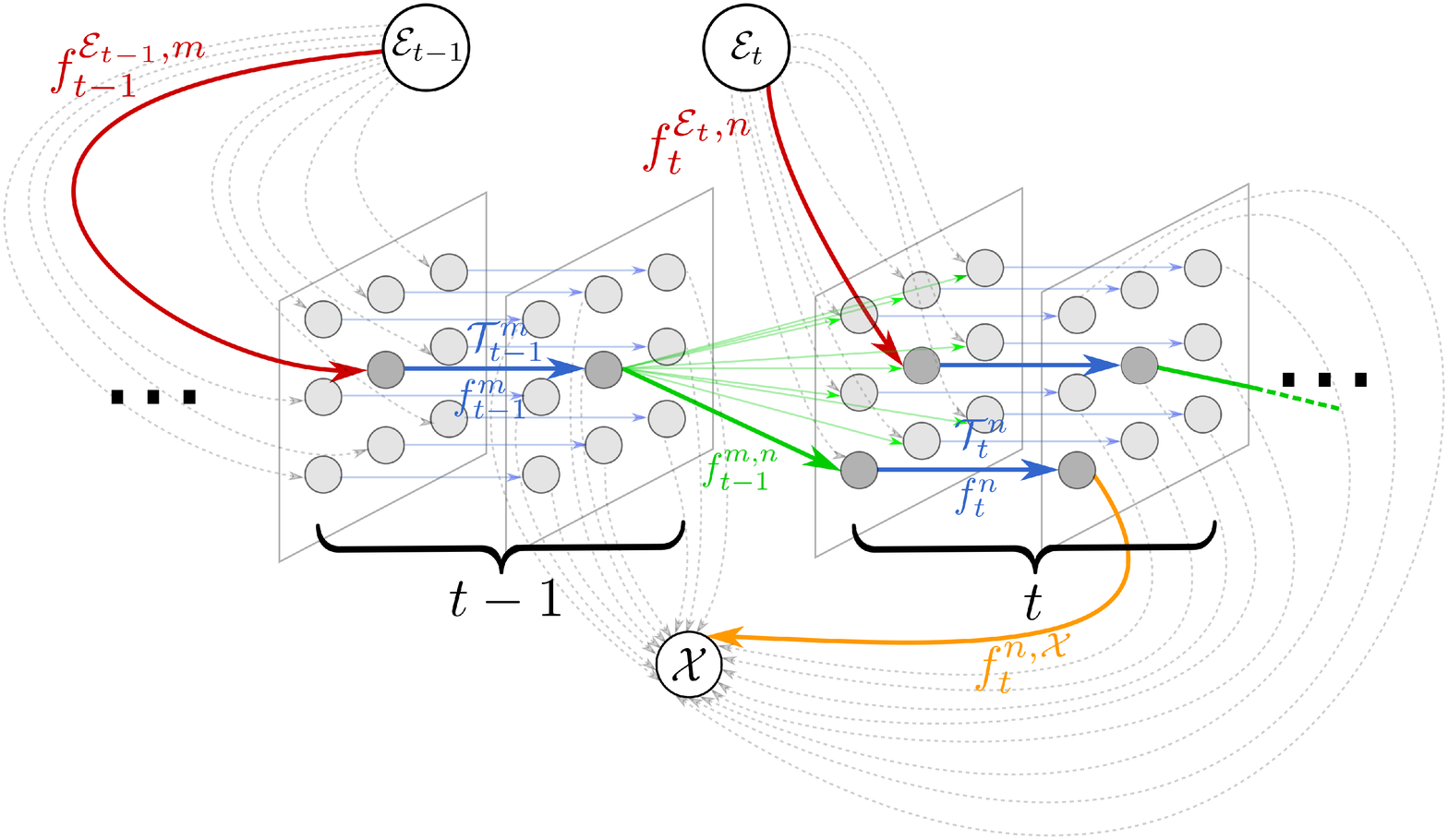}
\caption{Max-Flow graph (forward case). At each time frame $t$, a ``pseudo'' source node $\mathcal{E}_t$ is connected via an edge with flow $f_{t}^{\mathcal{E},n}$ (red) to tracklet $\mathcal{T}_t^n$. Each tracklet incurs a flow $f_t^n$ (blue) to pass through a superpixel $s_t^n$. Tracklets in frame $t$ are connected to tracklets in the next frame and allow for flows $f_{t}^{m,n}$ (green). The flow $f_{t}^{n,\mathcal{X}}$ can leave any tracklet in the network (orange).}
\label{fig:mcf}
\end{figure}

Thanks to its inherent structure, our MAP problem can be mapped into a cost-flow problem and can be solved efficiently. Specifically, we wish to determine where flow emitted from a source node must traverse a graph in order to minimize the traversal cost to a sink node~\citep{zhang08}. For that matter, we associate to each superpixel $s_t^n$ a tracklet $\mathcal{T}_t^n$ (\ie an edge that represents the entrance and exit of a superpixel) and define $r_t^n\in \mathbb{R}^2$ as the central pixel of superpixel $s_t^n$. Fig.~\ref{fig:mcf} shows a graphical representation of our flow network formulation.

As a first step, the MAP problem of Eq.~\eqref{eq:map2} is transformed into an Integer Program (IP)~\citep{schrijver98}. To simplify notations, let $\alpha_t^{m,n} \coloneqq P(Y_t^n=1|a_{t-1}^m)$, $\beta_t^n \coloneqq P(Y_t^n=1|a_t,g_t)$, and $\rho_t^n \coloneqq P(Y_t^n=1|\bm{a},\bm{g})$. We also introduce a sink node $\mathcal{X}$, and set of source nodes $\mathcal{E}_t$ such that each pushes flow onto the corresponding frame $I_t$.
Additionally, we introduce the variables $f_t^n$, $f_t^{m,n}$, $f_t^{\mathcal{E},n}$, and $f^{n,\mathcal{X}}$ to denote tracklet, transition, entrance and exit flows, respectively. The corresponding IP is thus given by,
  \begin{subequations}
  \label{eq:int_prog}
  \begin{align}
  \intertext{Maximize}
  &\sum_{t,n} \log{\frac{\rho_t^n}{1-\rho_t^n}}f_t^n + \sum_{t,m} \log{\frac{\alpha_t^{m,n}}{1-\alpha_t^{m,n}}}\sum_{t,n}f_t^{m,n} + \sum_{t,n} \log{\frac{\beta_t^n}{1-\beta_t^n}}f_t^{\mathcal{E}_t,n},\label{eq:loglikelihood}\\
  \intertext{subject to,}
  &\sum\limits_{n}f_t^{m,n} \leq 1, \qquad \forall t,m,n \label{eq:cap1_trans}\\
  &\sum_m f_t^{m,n} - \sum_p f_{t-1}^{p,m} \leq 0, \qquad \forall t,m,n,p \label{eq:conserv1}\\
  &\sum_{m,t} f_t^{\mathcal{E}_t,m} - \sum_p f^{p,\mathcal{X}} \leq 0, \qquad \forall t,m\label{eq:conserv2}
  \end{align}
  \end{subequations}
\noindent
where the above objective function, Eq.~\eqref{eq:loglikelihood}, corresponds to the log-likelihood of Eq.~\eqref{eq:map2} and where each flow variable associated to a cost term corresponds to a Bernoulli variable. The constraint defined by Eq.~\eqref{eq:cap1_trans} imposes a maximum flow capacitance of value one, thereby expressing the assumption that a superpixel can only contain a single target. Eq.~\eqref{eq:conserv1} imposes flow conservation, \ie at each node the input flow must be equal to the output flow (except for the source and sink nodes). Last, Eq.~\eqref{eq:conserv2} imposes that the sum of flow emitted by the source node $\mathcal{E}$ must reach the sink node $\mathcal{X}$. By design, the solution of this IP gives $Y_t^n=1$ if $f_t^n$ is equal to the edge capacity and 0 otherwise.

To further specify constraints for our given application, we now outline three additional edge-pruning measures. While these could be added directly in Eq.~\eqref{eq:loglikelihood}, we opt to describe them here instead.
\begin{itemize}
\item[-] {\bf Entrance edge pruning:} The provided 2D locations allow for a strong prior on the location of the object. Intuitively, we wish to force flow where the 2D locations are known (\ie $g_t$). We therefore connect $\mathcal{E}_t$ to all $\mathcal{T}_t^n$ such that the centroid of the corresponding superpixel, $r_t^n$, is included in a neighborhood centered at $g_t$ with radius $R$ (see Fig.~\ref{fig:temporal_merge}(left)). 
The parameter $R$ therefore control the quantity of flow that can be pushed from a given source node into its corresponding image. Edges that do not fulfil this condition are pruned.

\item[-] {\bf Transition edge pruning:} For edges that link tracklets, we use location and motion constraints to remove edges. For locations, we prune edges where $r_{t}^n$ is outside of a neighborhood centered on $r_{t-1}^m$ of radius $R$. Similarly, we estimate  superpixel motion by means of a histogram of oriented optical flow~\citep{chaudhry09}. Defining this motion by $u_t^n\in \mathbb{R}^l$, we prune edges such that $S_m(u_{t-1}^m,u_t^n) < \tau_u$, where $S_m(\cdot,\cdot)$ is the histogram intersection similarity.

\item[-] {\bf Tracklet edge pruning:} We let the probabilistic estimation described in Sec.~\ref{sec:foreground_model} be related to the cost of pushing flow through tracklets. Depending on the sequence, this estimate can lead to false positives (\ie give a high probability value on the background). To circumvent this phenomenon, we prune tracklet edges whose probability are below a threshold, $\rho_t^n < \tau_\rho$.
\end{itemize}

Note that given this IP formulation, the number of pseudo source nodes $\mathcal{E}_t$ does not change the optimization problem of Eq.~\eqref{eq:loglikelihood}. This implies that whether multiple 2D locations or none were be specified on each frame $t$, the IP would remain unchanged. Naturally, omitted source nodes on frames would reduce the quality of the solution as less information would be available to the foreground model (sec.~\ref{sec:foreground_model}). However, as we will show in our experiments, our global optimization recovers paths that span several frames and limits the impact of such cases. Similarly, the pruning of edges does not affect the solution of Eq.~\eqref{eq:loglikelihood} given that these would have infinite cost were they to be explicitly kept, and thus never allow flow to pass through them. 

\subsection{K-shortest path optimization}
As with all IP optimization problems, Eq.~\eqref{eq:int_prog} is NP-hard~\citep{papadimitriou81}. However, as noted in~\cite{berclaz11}, our problem can be relaxed to a Linear Program thanks to the total unimodularity of the constraints matrix. The latter condition guarantees that the solution will converge to an integer solution, making off-the-shelf optimizers suitable (\eg Simplex~\citep{klee70}, Interior point~\citep{kojima89}). However, we use a more efficient alternative -- the K-shortest paths algorithm (KSP) applied to the case where all edges have unit capacitance.
In contrast with generic LP solvers, KSP explicitly leverages the connectivity of nodes in the graph.
While~\cite{berclaz11} used a node-disjoint optimization to restrict nodes from receiving flow from different sources, our tracklet costs, $C_t^n$, allow a simpler edge-disjoint K-shortest paths algorithm by minimizing the negative of Eq.~\eqref{eq:loglikelihood}.
We provide further details on our implementation in Appendix A.

Last, to take into account information from previous and future frames, we compute both a forward and backward graph so to track superpixels forward and backward in time. This gives rise to two independent MAP problems to solve: one in each time direction. While we only present the forward case here, the backward case can easily be derived. The final labeling of a sequence is then given by the union of the two solution sets.

\subsection{Model costs}\label{sec:costs}
In what follows, we describe in detail how edge costs associated to Eq.~\eqref{eq:loglikelihood} are computed.

\begin{itemize}
\item[-]{{\bf Tracklet costs:}} As indicated in Sec.~\ref{sec:solving}, $\rho_{t}^n$ is the probability that the superpixel $s_t^n$ is part of the object according to the classifier. The cost of the corresponding flow $f_t^n$ is thus given by
\begin{equation}
C_{t}^n = -\log \frac{\rho_t^n}{1-\rho_t^n}
\label{eq:in_frame_cost}
\end{equation}
\noindent
and is illustrated with blue edges in Fig.~\ref{fig:mcf}. 

\item[-]{{\bf Transition costs:}} We model $\alpha_t^{m,n}$, the likelihood that superpixels $s_t^n$ and $s_{t+1}^m$ correspond to the same region in the sequence. In our flow-network, this corresponds to the cost of transiting from tracklet $\mathcal{T}_t^n$ to $\mathcal{T}_{t+1}^m$ (green edges in Fig.~\ref{fig:mcf}). 

While defining costs based on image features for such transitions is complex when the object size and background is unknown, we propose to learn and use an appropriate representation instead. In particular, we use Local Fisher Discriminant Analysis (LFDA), a supervised metric learning method~\citep{sugiyama06}, to measure the appearance similarity between two superpixels. In addition to Fisher Discriminant Analysis (FDA)~\citep{welling05}, which maximizes between-class scatter while minimizing within-class scatter, LFDA considers multi-modal classes, thereby preserving the local structure of data. In practice, we set the two LFDA parameters empirically: The $k$ nearest-neighbour data points used to compute an affinity matrix, and $Q$, the dimension of the output space. We select as positive samples the $a_t^n$ with associated probability $\rho_t^n$ larger than a threshold $\tau_{trans}$. As negatives, we randomly select an equal amount of samples below that threshold. Letting $V$ be the LFDA projection matrix, we set
\begin{equation}
  \alpha_{t}^{m,n} = \exp \left( {-|| V(a_t^n - a_{t+1}^{m})||^2_2} \right),
\label{eq:alpha}
\end{equation}
\noindent
and the flow cost $f_t^{m,n}$ is then given by
\begin{equation}
C_{t}^{m,n} = -\log\frac{\alpha_{t}^{m,n}}{1-\alpha_{t}^{m,n}}.
\end{equation}

\item[-]{{\bf Entrance-Exit costs:}}
The source nodes $\mathcal{E}_t$ allow flow to be pushed through the network. Intuitively, at a given frame, we want to push flow starting from superpixels that are similar to the region given by$g_t$. Letting $\beta_t^n = P(Y_t^{n}|a_t,{g}_{t})$ denote the probability of entering the network from $\mathcal{T}_t^n$, we compute,
\begin{equation}
  \beta_{t}^n = \exp \left( -|| V(a_t^n - a_t)||^2_2 \right)
\label{eq:beta}
\end{equation}
where $V$ is the previously computed LFDA projection matrix, $a_t$ corresponds to the feature vector of the superpixel selected by $g_t$, whereas $a_t^n$ corresponds to the superpixel feature vector in question. The flow cost associated $f_{t}^{\mathcal{E}_t,n}$ (red edges in Fig.~\ref{fig:mcf}) is then taken as $C_{t}^{\mathcal{E}_t,n} = -\log(\beta_{t}^n / (1-\beta_{t}^n))$. In addition, since we do not model the likelihood of terminating a path, we set the cost of exiting to the sink node $C_{t}^{n,\mathcal{X}}$ to be 0 for all tracklets.
\end{itemize}

\begin{figure}[t]
\centering
\includegraphics[width=0.49\textwidth]{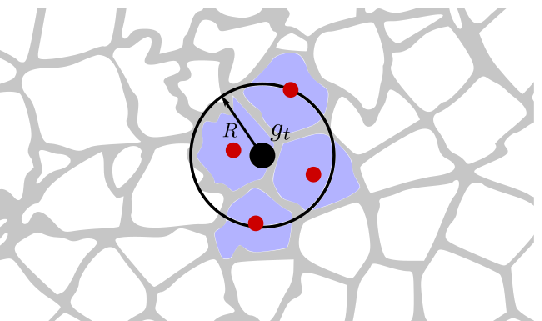}
\includegraphics[width=0.49\textwidth]{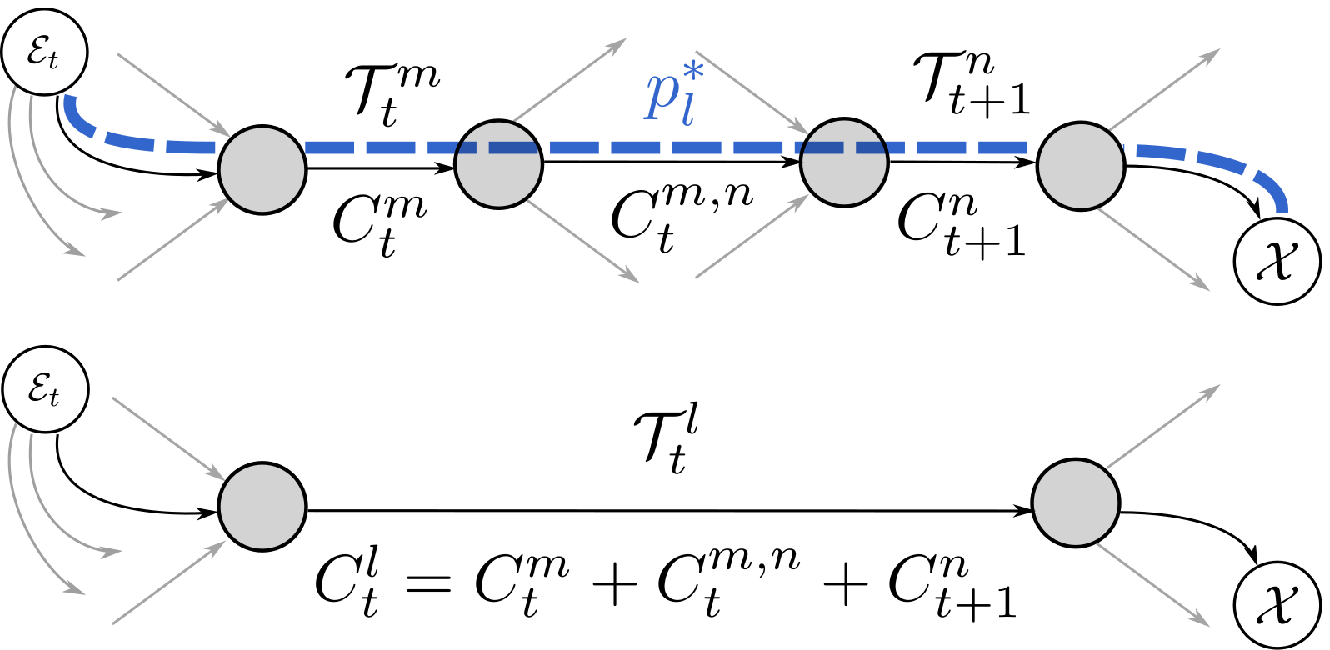}
\caption{(left) Flow entrance. Superpixel boundaries are drawn in grey. All superpixels whose centroids (red circles) are contained in a circle of radius $R$ will accept entrance flow. (Right) Example of a temporal merge. Top: Path $p_l^*$ (in blue) crosses tracklets $\mathcal{T}_t^n$ and $\mathcal{T}_{t+1}^{m}$. Bottom: At the next iteration, the corresponding tracklets merge into a single tracklet $\mathcal{T}_t^{l}$ with cost $C_t^n + C_{t}^{m,n} + C_{t+1}^{m}$.}
\label{fig:temporal_merge}
\end{figure}


\subsection{Iterative tracking}
\label{sec:iterative_ksp}
Given that our object model described in Sec.~\ref{sec:foreground_model}, is trained on very few positive samples, solving Eq.~\eqref{eq:loglikelihood} can lead to a number of missed positive superpixels. To circumvent this limitation, we propose to augment the positive set $\mathcal{S}_p$ in an iterative way, by adding new positive samples recovered from our produced KSP estimation. That is, after initially inferring the object segmentation, new object superpixels are considered as positive samples in order to re-train the classifer and recompute the graph costs. 

To do this, the costs $C_t^n$, $C_{t}^{m,n}$ and $C_{t}^{\mathcal{E},n}$ are updated after each KSP optimization. More specifically, we define $\mathcal{P^*} = \{ p_0^*,...,p_{K-1}^*\}$ to be the set of solution paths given by the KSP optimization, with $p_l^*$ being the set of tracklets in path $l$. We make the assumption that at the next iteration, the solver would most likely extend found paths given by the previous result. We can then merge tracklets belonging to the same path (\ie concatenate them temporally to form a new tracklet). This brings the practical advantage of reducing the complexity of our problem as the number of edges decreases at each iteration. 

\begin{figure}[t]
\centering
\includegraphics[width=0.99\textwidth]{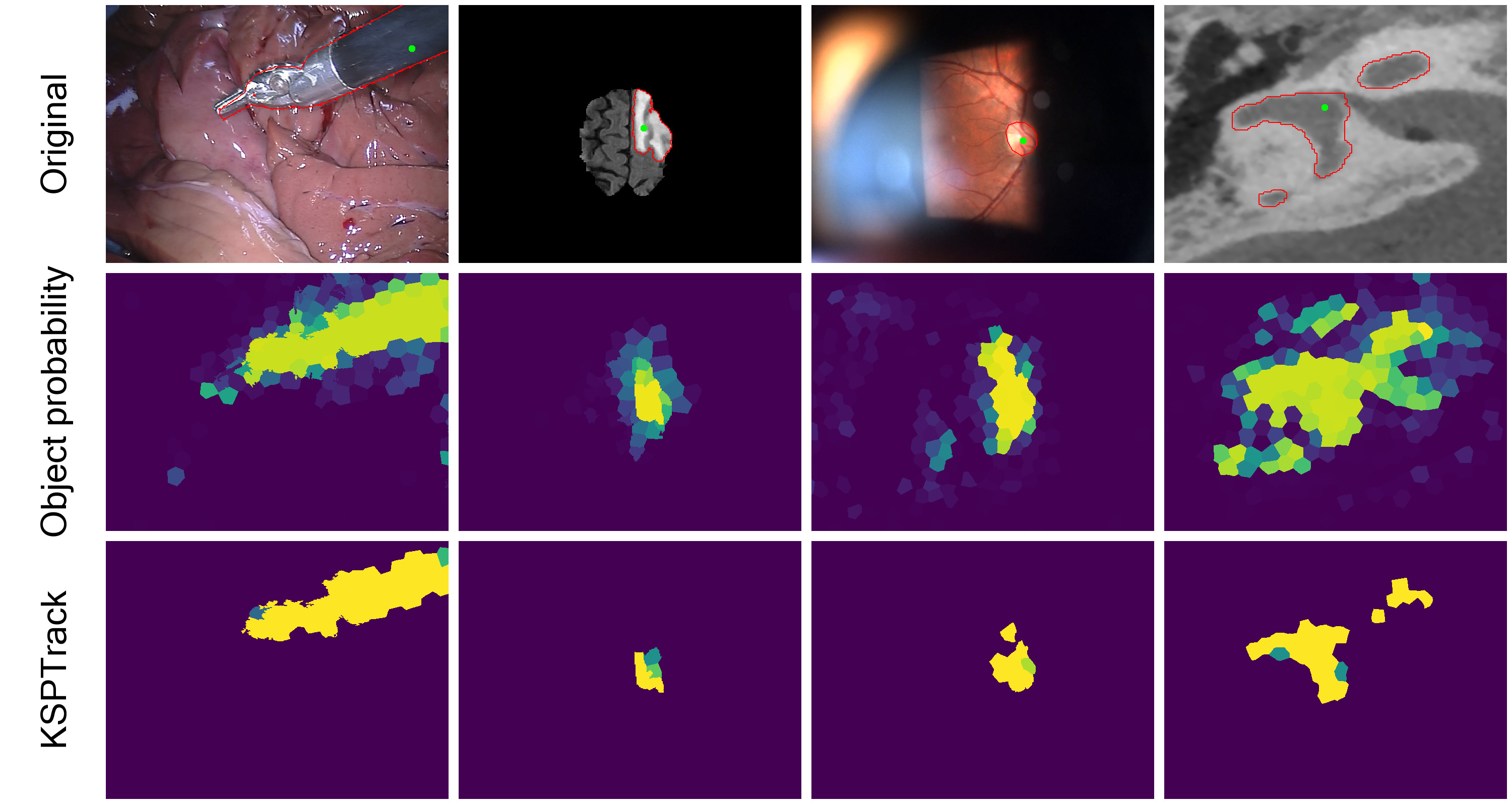}
\caption{(Top row): Original images from different datasets. Ground truth contour of the structure of interest is depicted in red and the supervised 2D locations are shown in green.
(Middle row): $\rho_{t}^n$, probability estimates of the object given by our classifier after the final iteration of our approach.
(Bottom row): Pixel-wise sum of binary segmentations after each iteration of the KSP optimization. Total number of iterations from left to right are: $3$, $4$, $3$, $2$.}
\label{fig:example_iter_ksp}
\end{figure}

In this case, we set the edge cost following the merge to be
\begin{equation}
\begin{split}
  C_l := \sum_{n,t}C_t^n + \sum_{m,n,t}C_{t}^{m,n}
\end{split}
\label{eq:cost_transform}
\end{equation}
\noindent
with tuples $(n,t)$ and $(m,n,t)$ corresponding to edges occupied by path $p_l^*$. The algorithm then terminates when no new tracklets are added to the set $\mathcal{P}$. Fig.~\ref{fig:temporal_merge}(right) illustrates this temporal merging step while the pseudo code of our iterative solver, which we denote \KSP, is shown in Alg.~\ref{alg:iter_ksp}. Fig.~\ref{fig:example_iter_ksp} shows example frames of how different samples are sequentially added to the positive set, allowing for better classification and KSP solutions.

\begin{algorithm}[t]
\SetKwInOut{Input}{Input}
\SetKwInOut{Output}{Output}
\DontPrintSemicolon
\Input{$\mathcal{S}_p$: Initial set of positive superpixels, $\mathcal{S}_u$: Initial set of unlabeled superpixels, $\mathcal{S}$: set of superpixels (with associated variables $\bm{a}$, $\bm{u}$, and $\bm{r}$), $\mathcal{G}$: 2D locations}
\Output{$\mathcal{P}$: Set of K-shortest paths}
$g \gets$ \texttt{make\_graph}$(\mathcal{G},\mathcal{S}_p,\mathcal{S}_u,\mathcal{S})$\tcp*{As in sections \ref{sec:foreground_model} and \ref{sec:costs}}
$\mathcal{P} \gets \emptyset$ \tcp*{Initialize output set to null}
$find\_paths \gets$ True\tcp*{Flag variable (disabled at convergence)}
\While{$find\_paths$}{
  $\mathcal{P^*} \gets$ \texttt{run\_k\_shortest\_paths}$(g)$\tcp*{As in ~\ref{sec:ksp}}
 \eIf(\tcp*[f]{$\phi(.)$ gives the quantity of superpixels}){$\phi(\mathcal{P}) = \phi(\mathcal{P^*})$}{
   $find\_paths \gets false$\;
 }{
   $g \gets$ \texttt{update\_tracklet\_costs}$(g,\mathcal{P^*})$\tcp*{Update $\mathcal{S}_p$ and do as in Sec.~\ref{sec:foreground_model}}
   $g \gets$ \texttt{update\_entrance\_transition\_costs}$(g,\mathcal{P^*})$\tcp*{As in sections \ref{sec:costs}}
   $g \gets$ \texttt{temporal\_merge}$(g,\mathcal{P^*})$\tcp*{As in Sec.~\ref{sec:iterative_ksp}}
   }
   $\mathcal{P} \gets \mathcal{P^*}$\;
   }
\caption{\KSP ~Algorithm (single direction).\label{alg:iter_ksp}}
\end{algorithm}

\section{Experiments}
\label{sec:experiments}
The following section details the implementation of our approach, as well as the parameter values used. We then outline the image datasets that we evaluate and the baselines methods used to compare performances.

\subsection{Implementation and Computational cost}
Our \KSP ~method is implemented in Python/C++\footnote{We make our implementation, along with the tested datasets and corresponding manual ground truth segmentations available at \label{fn:website}\href{www.gazelabel.com}{www.gazelabel.com}}. Using a Linux machine equipped with a Quad-core 3.2 GHz Intel CPU, the construction of both the forward and backward graphs take 15 minutes each. Superpixel segmentation, including the extraction of dense optical flow, requires 10 minutes. Each iteration in Alg.~\ref{alg:iter_ksp} takes 5 minutes, including the training of the classifier and computing the entrance-exit models. The number of KSP iterations varies between 1 and 5 depending on the sequence and the provided 2D locations. A GeForce GTX 1080 Ti GPU and a Keras based implementation of our network was used for our IOS features, taking 3 hours for 20 epochs (at 500 iterations per epoch).
In total, our method therefore takes roughly 4 hours for a sequence of 120 frames.

\subsection{Selection of Parameters}
Table~\ref{tab:parameters} specifies the values of the parameters used in the experiments that follow. Note that these are fixed once and for all, over all experiments and for all datasets. These values were selected empirically so to perform well over all tested sequences.

\begin{table}[htbp]
\centering
\begin{tabular}{c p{8cm} c }
\toprule
Symbol & Description & Value \\
\toprule
$N_t$   & Approximate number of superpixels per frame & $520$\\
$M$   & Number of trees of bagging classifier& $500$ \\
$\tau_{\rho}$   & Threshold on probabilities of foreground model & $0.5$\\ 
$\tau_{u}$   & Threshold on histogram intersection cost & $0.75$ \\ 
$\tau_{trans}$  & Threshold on appearance-transition probabilities & $0.9$\\
$k$ & Number of clusters for LFDA & $5$\\ 
$D$ & Number of dimensions for LFDA & $7$\\ 
$R$   &Normalized radius of entrance/transition neighborhood & $0.05$ \\ 
$\sigma_{g}$  & Normalized standard-deviation for prior in feature extraction & $0.3$ \\
\bottomrule
\end{tabular}
\caption{Summary of the parameters used in \KSP.}
\label{tab:parameters}
\end{table}

\subsection{Datasets}
\label{sec:data}
We evaluate our method on a mixture of datasets consisting of video sequences and volumetric images. Note that the datasets include a variety of different image modalities with a wide range of applications. The singularities of each sequence are given so as to emphasize the flexibility of our method. Note that for each sequence and for all datasets, a single object of interest is present on any give image:

\begin{itemize}
\item[-]{\bf{Brain:}} $4$ randomly selected volumetric sequences from the publicly available BRATS challenge dataset~\citep{menze15}. Each volume contains a 3D  T2-weighted MRI scans of a brains containing a tumor, which we choose as the object of interest. The tested volumes contain ${73,69,75,74}$ slices each of size ($240 \times 240$). Tumors have in general a ball-like shape, \ie their radius increases and decreases as slices unfold.

\item[-]{\bf{Tweezer:}} We extract $4$ sequences from the training set of the publicly dataset MICCAI Endoscopic Vision challenge: Robotic Instruments segmentation~\citep{endochal}. Each extracted sequence contains $121$ frames and are acquired at $25$ fps. The object to segment in each sequence is a surgical instrument, and where each frame is of size ($640 \times 480$). The tool is piecewise-rigid and is subject to translations and rotations in an otherwise stable environment.
  
\item[-]{\bf{Slitlamp:}} $4$ slit-lamp video recordings of human retinas, where the optic disk must be segmented. The sequences contain ${129,121,75,130}$ frames of size ($680 \times 512$), all acquired at 25 fps. The object has a relatively constant shape and texture, but  undergoes abrupt translations occasionally. Due to this imaging technique, the background also changes lighting abruptly, with non-global bright beams of yellow and blue light appearing.

\item[-]{\bf{Cochlea:}} $4$ volumes of 3D CT scans of the inner ear, where the cochlea must be annotated. The challenge with this dataset lies in the fact that the object of interest branches out in several parts and merges back. Volumes contain ${99,96,116,104}$ slices of size ($300 \times 290$).
\end{itemize}

For the {\bf Slitlamp} and {\bf{Cochlea}} datasets, we manually segmented the object ground truth on each frame in each image sequence. Manual pixel-wise annotations are publicly available for the {\bf Brain} and {\bf Tweezer} datasets. The complete set of used image sequences and manual ground truths are publicly available \footref{fn:website}.

\subsection{Generating 2D coordinate locations}
Unless otherwise specified, 2D coordinate locations, $g_t$ for each sequence were collected using an off-the-shelf gaze tracker (Eye Tribe Tracker, Copenhagen, Denmark) as in~\cite{lejeune17}. To do this,  the tracker was placed beneath a $12.3"$ tablet roughly $50cm$ away from a user's face. For each recording session, an initial calibration procedure was performed using the inbuilt software of the tracker and validated before all gaze information recordings took place, allowing less than 1$^{\circ}$ tracking accuracy at 30fps.

Gaze recordings were then collected by a domain expert who had been instructed to observe the object of interest throughout the sequence. During the displaying of the video, gaze locations were recorded using a dedicated software\footref{fn:website}. Videos were displayed at $10$ fps and 2D locations were then taken to be the average ($x,y$) coordinates given by the tracker over the corresponding time interval. As such, excluding the initial calibration phase, annotating a 100 frame sequence with a single 2D location per frame took roughly 10 seconds. 

\subsection{Baselines}
To compare our approach to existing methods in the literature, we evaluate the following closest methods:

\begin{itemize}
\item[-]{\PSVM :} Patch-based SVM
is a transductive learning approach~\citep{vilarino07} explicitly developed to use gaze information to produce \blue{segmentations} when viewing endoscopy video sequences. 

\item[-]{\GS :} This approach used gaze trackers to annotate CT volumes using a saliency map-construction and a Random-Walker to segment the object of interest~\citep{khosravan16}. 

\item[-]{\EEL :} An expected exponential loss was proposed to learn robust classifiers in a PU learning setting~\citep{lejeune17}. As in this paper, the method is presented over a variety of image datasets and used a gaze tracker to specify 2D coordinates. 

\item[-]{\DL :} Point location supervision was used to train a CNN while using a strong object prior to provide additional information to the network~\citep{bearman16}. The method was demonstrated to perform well on natural images.
\end{itemize}

To compare these methods, we implemented both \PSVM ~and \GS ~following their description, while we used provided code for \EEL ~and \DL . \PSVM, \GS, \EEL ~and \DL ~require approximately $8, 1, 3, 4.5$ hours respectively to process sequences. This computation time is roughly equivalent to our proposed method.

\section{Results}
\label{sec:results}
To validate our approach across a wide range of settings, we report the following 5 experiments: 
(i) The performance of the proposed method is compared to the baseline methods for all sequences in terms of segmentation accuracy; (ii) we compare the performance of a supervised prediction method when trained with ground truth generated by hand or with our approach; (iii) we assess the robustness of our method with respect to the selection of the 2D locations; (iv) we evaluate our IOS feature extraction strategy and compare it to different alternatives; (v) we assess the robustness of our method with respect to outliers and missing 2D locations.

\subsection{Experiment 1: Accuracy of produced segmentation}
\label{sec:accuracy}
\begin{figure}[t!]
    \centering
    \begin{subfigure}[b]{0.5\textwidth}
        \centering
        \includegraphics[height=3.5cm]{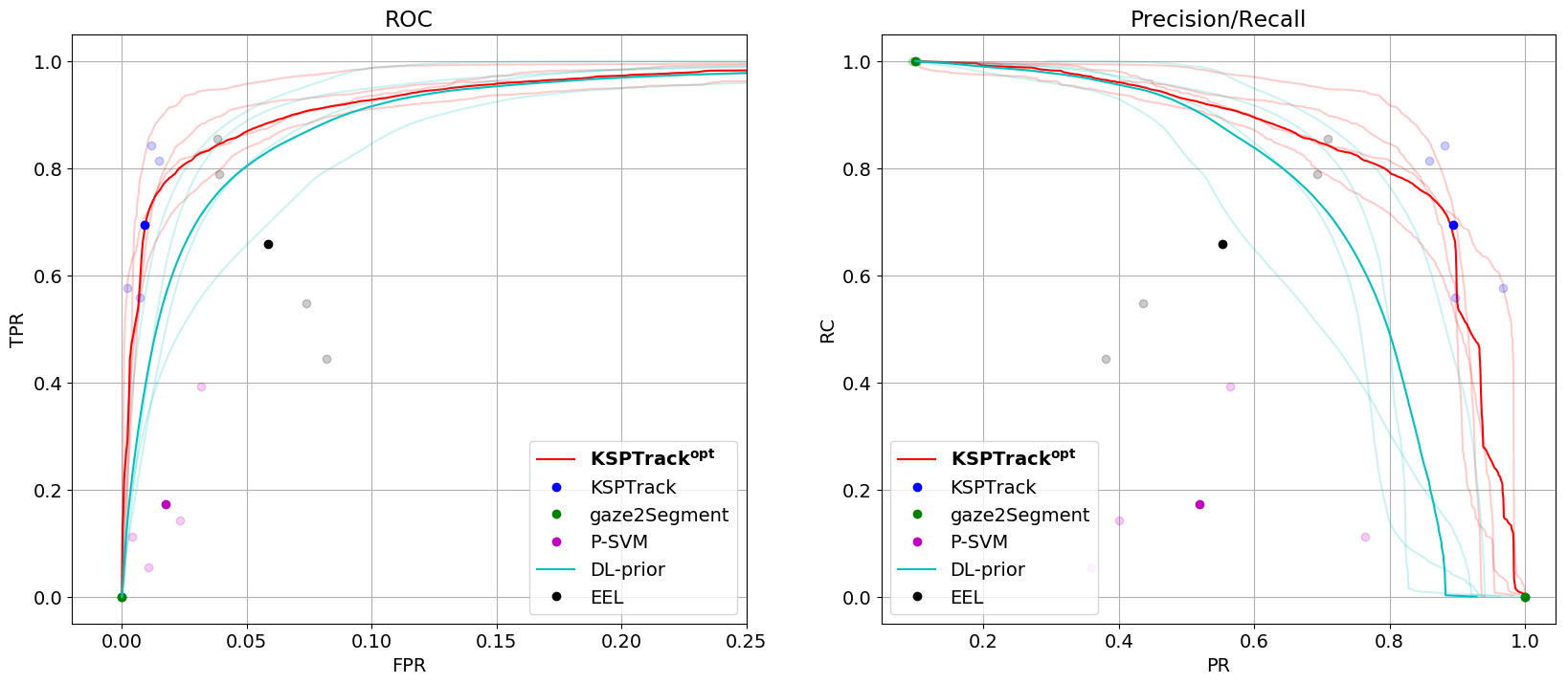}
        \caption{Tweezer}
    \end{subfigure}%
    ~ 
    \begin{subfigure}[b]{0.5\textwidth}
        \centering
        \includegraphics[height=3.5cm]{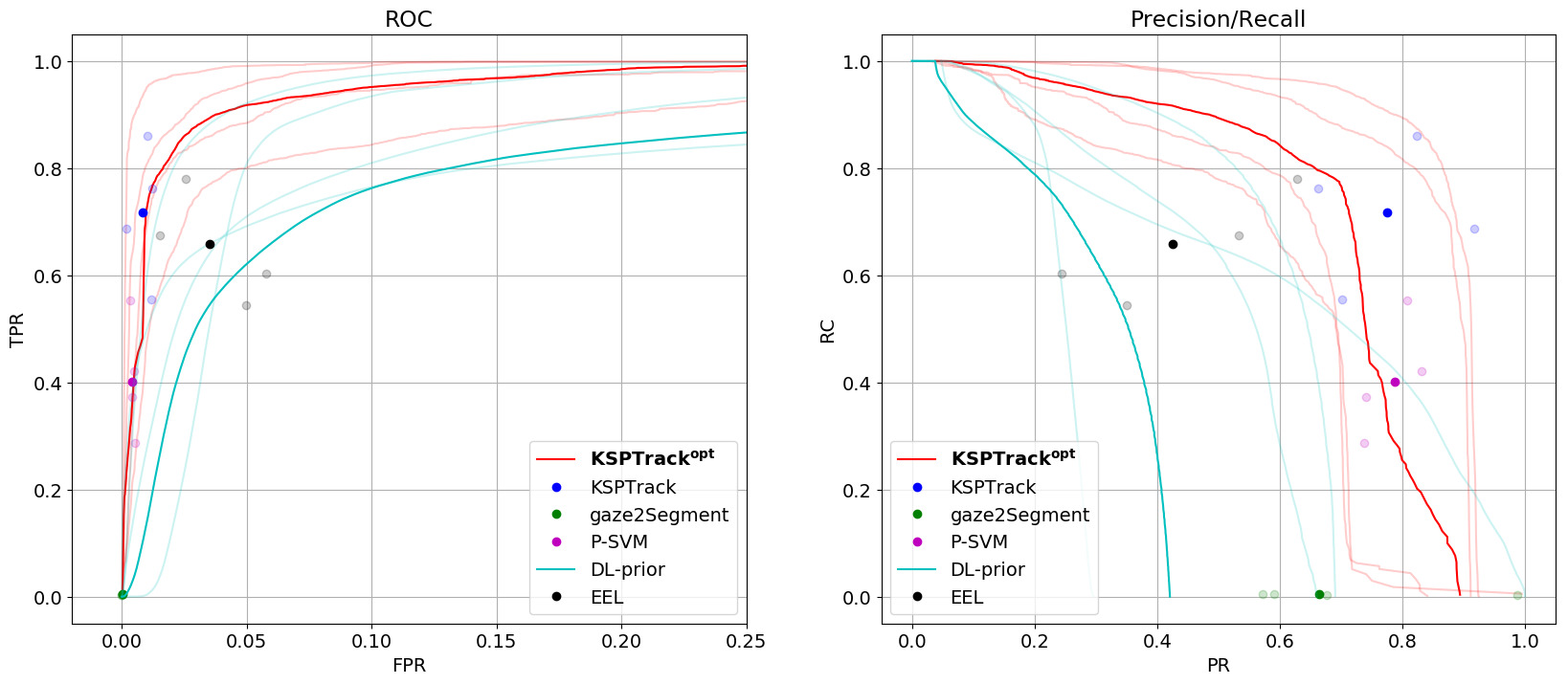}
        \caption{Brain}
    \end{subfigure}
    \begin{subfigure}[b]{0.5\textwidth}
        \centering
        \includegraphics[height=3.5cm]{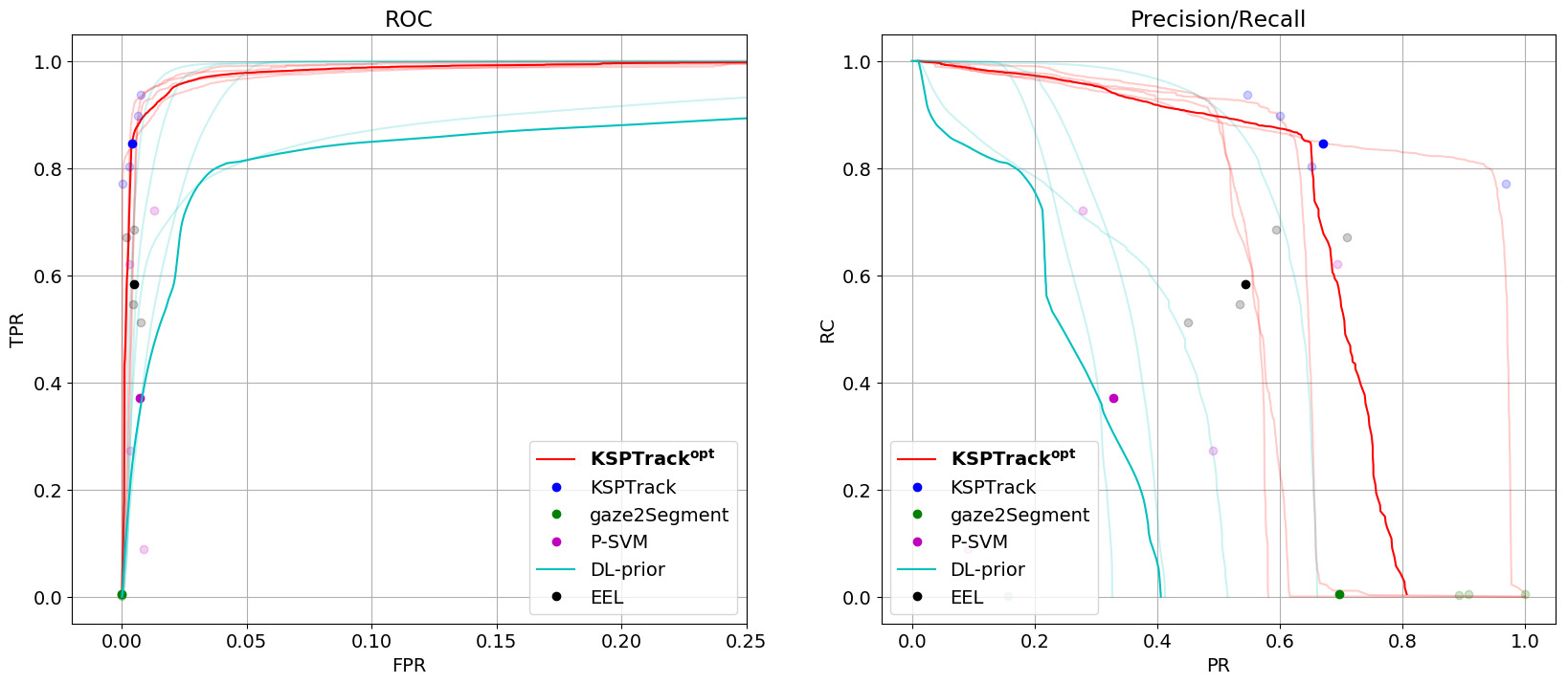}
        \caption{Slitlamp}
    \end{subfigure}%
    ~ 
    \begin{subfigure}[b]{0.5\textwidth}
        \centering
        \includegraphics[height=3.5cm]{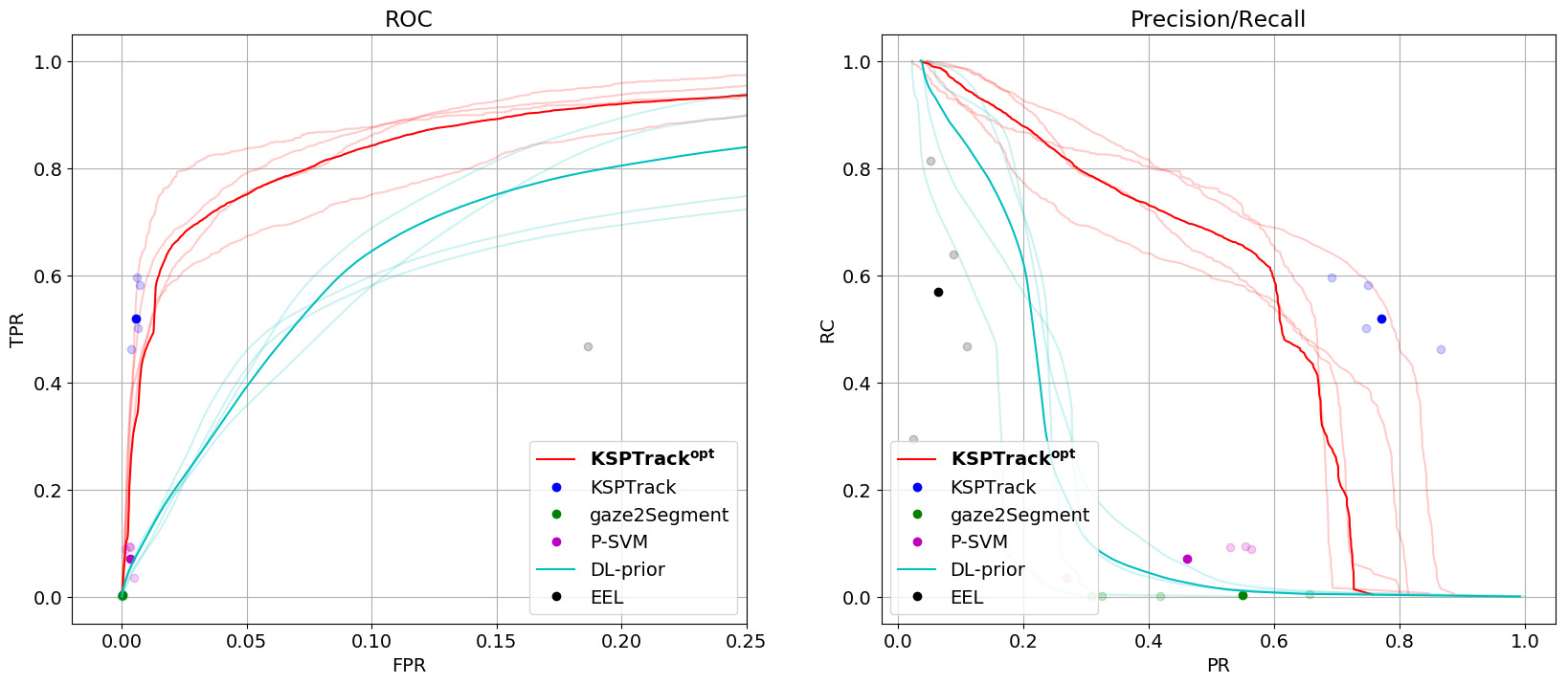}
        \caption{Cochlea}
    \end{subfigure}
    \caption{ROC and Precision-Recall curves for all types of sequence. In each case, we show the performance on each sequence (in light color) and averaged over each dataset (in bold)}
    \label{fig:ROC-PR}
\end{figure}

We first compare the accuracy of pixel-wise segmentations produced by our method and the baselines.
We illustrate in Fig.~\ref{fig:ROC-PR} the ROC and Precision-Recall curves for each method and for each dataset. In each case, we show the performance on each sequence (in light color) and averaged over each dataset (in bold). To measure segmentation accuracy, we also compute the F1-score for each method on each sequence and report these values in Table.~\ref{tab:stats}.

In addition, we distinguish our method in two: \KSP ~and \KSPOPT. In the former, we take the output of the method from the optimization directly. In the latter, we use the probabilities provided by our foreground model after training on the solution of \KSP. We then select the best threshold to maximum performance. As such, \KSPOPT ~can be viewed as the optimal segmentation one could hope for if we had a validation set, while \KSP ~uses no additional information to infer any threshold.

We report substantial improvement over all sequences in both categories. On the {\bf Tweezer} sequence, we improve the best baseline by
$196\%$ (\KSP ~vs.~\PSVM ~) and $12\%$ (\KSPOPT ~vs.~\DL). On the {\bf Brain} sequences, we improve by $40\%$ (\KSP ~vs.~\PSVM) and $36\%$ (\KSPOPT ~vs.~\DL).
On the {\bf Slitlamp} sequence, we report an improvement of $108\%$ (\KSP ~vs.~{\bf P-SVM}), and $32\%$ (\KSPOPT ~vs.~{\bf DL-prior}).
Similarly, for {\bf Cochlea}, we improve over the best baseline by $370\%$ (\KSP ~ vs.~\PSVM), and $113\%$ (\KSPOPT ~vs.~\DL).

As illustrated in Fig.~\ref{fig:all}, the results of our methods show improved segmentations compared to tested baselines from a qualitative point of view. To further depict the tracking that our approach produces, Fig.~\ref{fig:brain_paths} shows how the tracklet association of different superpixels across frames in a {\bf Brain} sequence for both the forward and backward passes of the optimization. Here the tumor is initially small (\ie frame 1), then grows (\ie frames 16, 31 and 46) to ultimately shrink again (\ie frame 61). We can see that certain superpixels are tracked over multiple frames even though the number of regions to segment varies across the frames.

\begin{figure}
\centering
\includegraphics[width=0.5\textwidth]{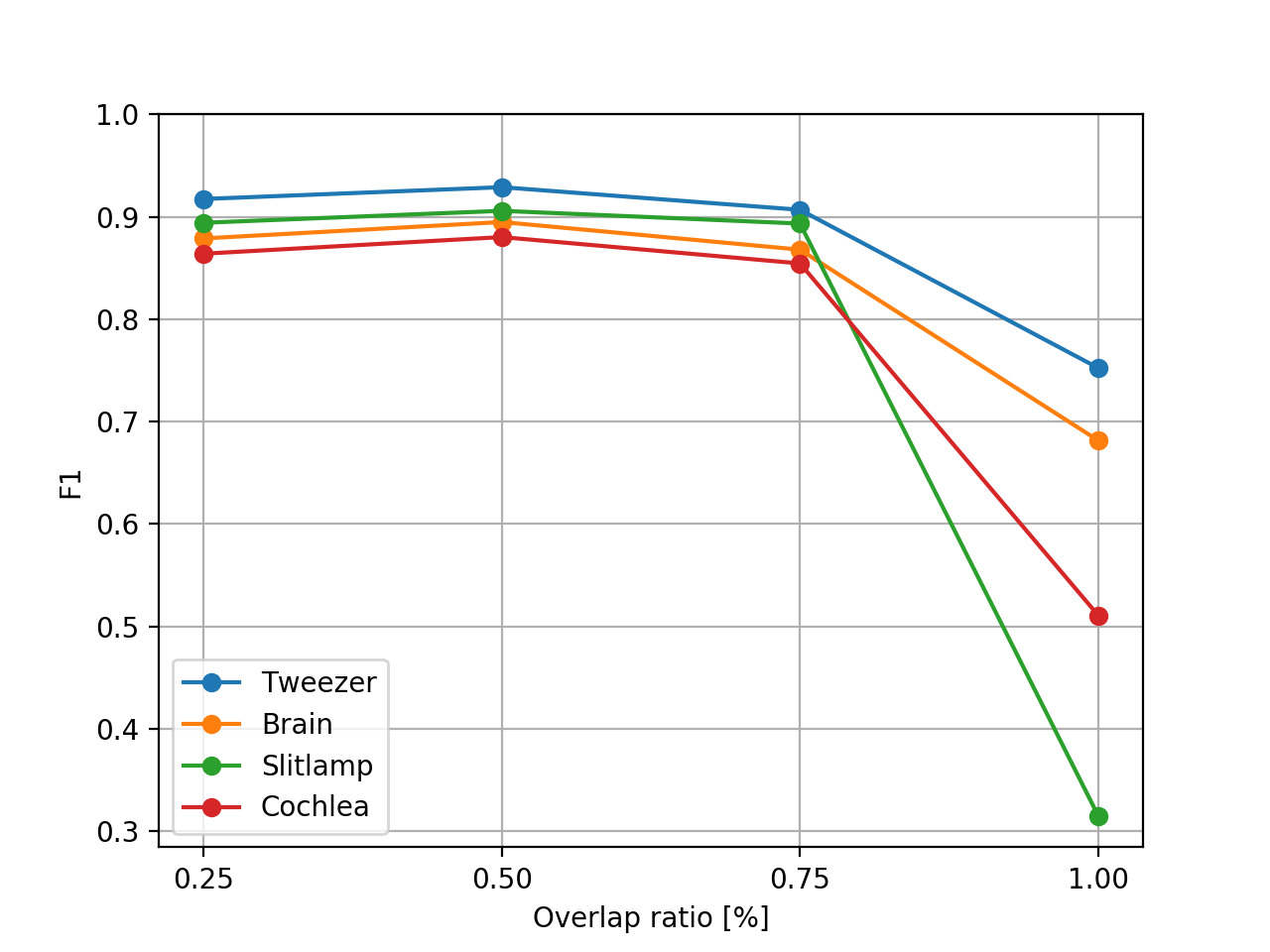}
\caption{\blue{Accuracy of superpixel segmentation (F1 score) for each dataset type when using different proportions of positive pixels within a superpixel to define a positive superpixel. F1 scores are averaged over 4 sequences.}}
\label{fig:sp_limits}
\end{figure}

\begin{figure}
\centering
\includegraphics[width=0.99\textwidth]{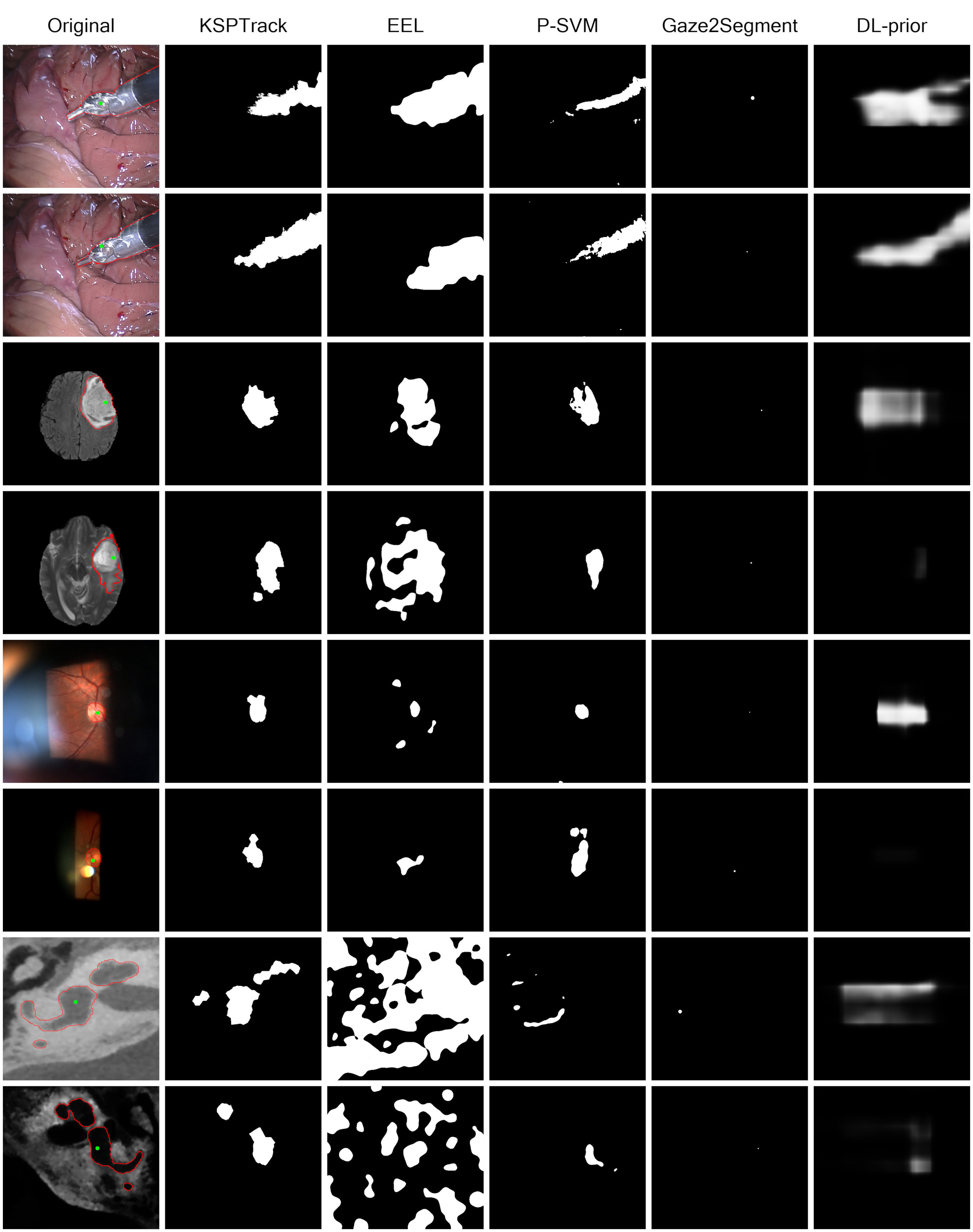}
\caption{Qualitative results of compared methods on the tested datasets. (First column) Original image. Ground truth contour of structure of interest is depicted in red and the 2D location is shown in green. (Second row onward) Binary segmentation of methods: \KSPnb ~(Proposed), \EELnb, \PSVMnb, \GSnb, and \DLnb.}
\label{fig:all}
\end{figure}

\begin{figure}[t]
\centering
\includegraphics[width=0.99\textwidth]{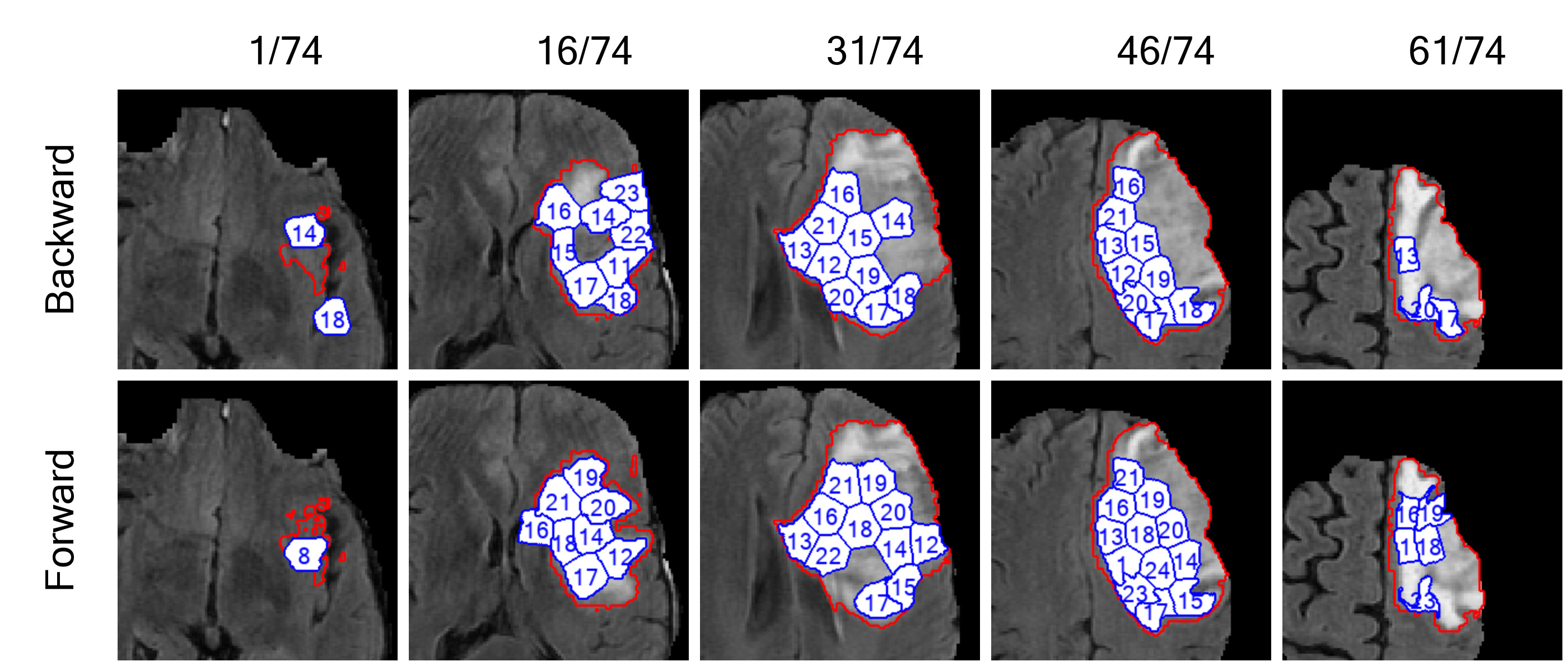}
\caption{
Example paths in the forward and backward tracking directions on a Brain sequence. The ground truth is highlighted in red. Segmented superpixels are highlighted in blue. Numbers within superpixel regions denote indices of paths.
}
\label{fig:brain_paths}
\end{figure}

\blue{
  Note that the performance of our approach is bounded by the quality of the superpixels used. In particular, some superpixels may contain both foreground and background pixels which reduces the best case performance of our approach. To quantify the impact of the superpixels on the produced segmentations, we were interested in looking at the F1 score if our approach  produced a ``perfect" labeling. To do this, we computed the F1 score between the manual ground truth and the set of positive superpixels when positive superpixels are defined by having more than a given proportion of positive pixels (\ie proportions of $0.25$, $0.5$, $0.75$ and $1$). That is, a proportion of $1$ is when all pixels in a superpixel are in fact positive pixels. Fig.~\ref{fig:sp_limits} illustrates the relation between these proportions and the average best case F1 score for each datasets. From this, we note that even if our method correctly labeled all superpixels, the F1 score would not be $1$. Also, if a strict threshold were to be used to denote positive superpixels (\ie $1$ or no negative pixels in a superpixel), our approach would provide near perfect performances.
}

\begin{table}
  \centering
  \begin{tabularx}{\textwidth}{c @{\extracolsep{\fill}} c cccc cccc}
      \toprule
      \multirow{2}{*}{Type} & \multirow{2}{*}{Method} & \multicolumn{4}{c}{F1} &
      & F1 & PR & RC \\
       & & 1 & 2 & 3 & 4 & & mean $\pm$ std& mean $\pm$ std& mean $\pm$ std\\ 
      \fatline
      \dataall{tables/Tweezer_self.csv}{Tweezer}
      \dataall{tables/Brain_self.csv}{Brain}
      \dataall{tables/Slitlamp_self.csv}{Slitlamp}
      \dataall{tables/Cochlea_self.csv}{Cochlea}
  \end{tabularx}
\caption{Comparison of quantitative results on all datasets. We report the F1 score for each method on each tested sequence using 4 different 2D gaze sets. In addition, for each sequence type, we give the mean and standard deviation F1, precision (PR) and recall (RC) scores.}
\label{tab:stats}
\end{table}

\subsection{Experiment 2: Segmentation performance when using generated \blue{segmentations}}
We now investigate the setting where one wishes to train a segmentation classifier to predict unseen sequences using either manually generated ground truths or \KSP ~produced \blue{segmentations}. That is, we wish to assess the bias of our method may induce by comparing the performance of a classifier that is trained with one type of \blue{segmentation}.

In our experiments, for each dataset, we use 3 out of the 4 sequences to train a standard U-Net using hand-annotated or \KSP ~produced \blue{pixel wise segmentation}. We use the binary cross-entropy as a loss function and perform a leave-one-out prediction (\ie train on $3$ sequences and predict on the last sequence). We keep as validation set $5\%$ of the frames belonging to the train sequence. Each model is trained for $40$ epochs at $500$ iterations per epoch. The model with the lowest validation loss is used in the prediction phase. We compute for each type and each fold the maximum F1-score obtained by both types of segmentations.

\begin{figure}[h]
\centering
\includegraphics[width=0.99\textwidth]{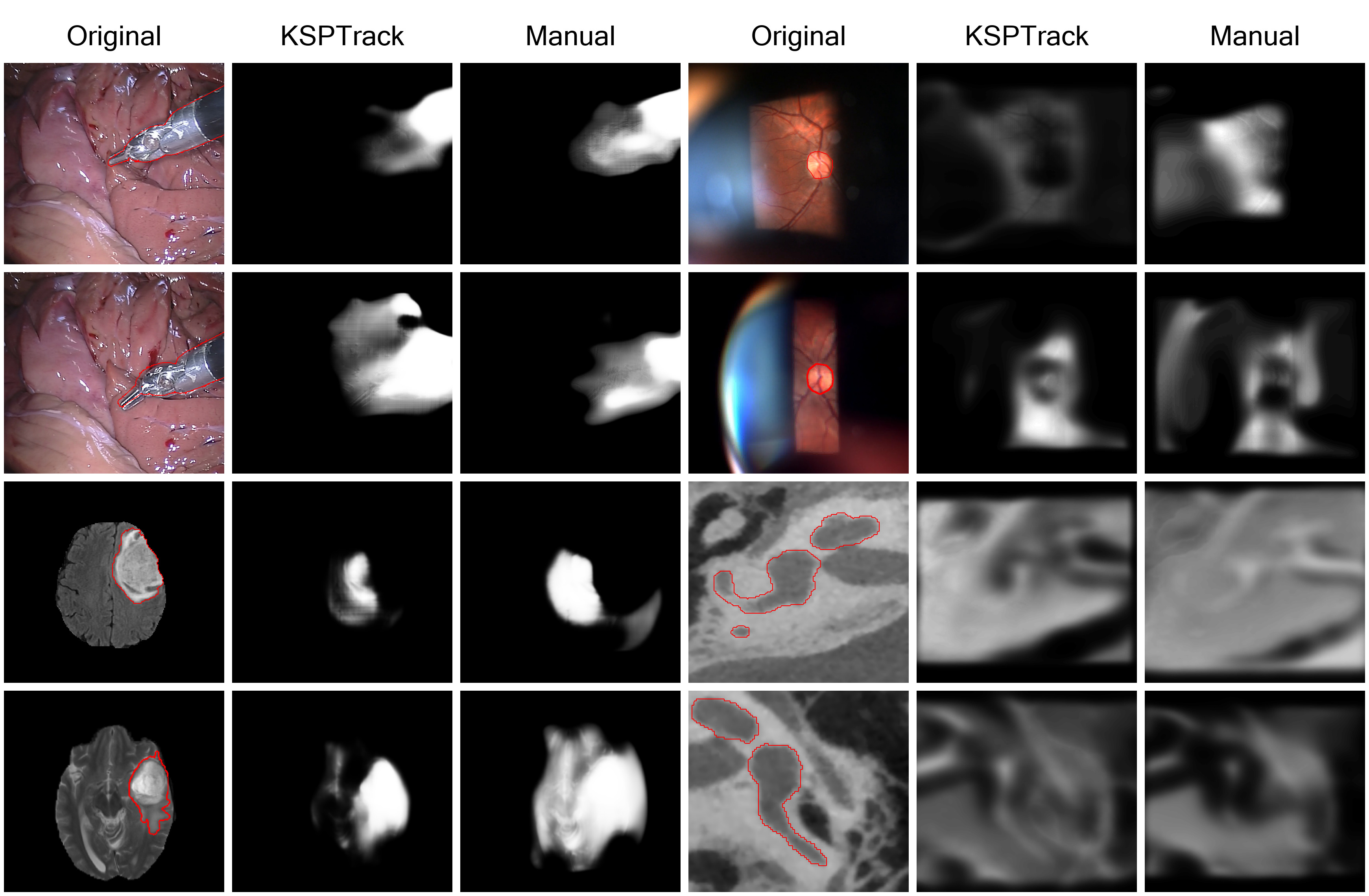}
\caption{Maximum F1 scores when using \KSPnb or manual segmentations to train a supervised CNN for pixel wise predictions.}
\label{fig:learning}
\end{figure}

Table~\ref{tab:learning} shows the mean scores over the 4 folds when using the hand and \KSP ~\blue{segmentations}, while Fig.~\ref{fig:learning} illustrates example predictions. In particular, we report a gap of $-10\%$ in prediction on the {\bf Tweezer} dataset. The {\bf Brain} sequence shows a gap of $-5\%$. For the {\bf Slitlamp} sequence, a gap of $-16\%$ is obtained. The {\bf Cochlea} sequence shows a gap of $-10\%$.

In general, the fact that our approach provides segmentations that are qualitatively inferior to their hand-based counterpart affects the performance of the classifier in the prediction setup. However, this performance decrease is not overwhelming and depends significantly on the variability of the sequences. In particular, the {\bf Slitlamp} datasets contains a wide range of sequences that appear different from one another. As such, it is not surprising that this dataset suffers the least when compared to the others. 
\begin{table}[h]
\centering
\begin{filecontents*}{tables/learning_all.csv}
,KSP,True
Tweezer,0.556,0.611
Brain,0.398,0.421
Slitlamp,0.077,0.092
Cochlea,0.112,0.124
\end{filecontents*}
\csvreader[no head,tabular=ccc,
table head=\\\toprule Types & \KSPnb & Manual\\\fatline,
table foot=\hline,
filter expr={
      test{\ifnumgreater{\thecsvinputline}{1}}
    }]
{tables/learning_all.csv}
  {1=\type,2=\sksp,3=\smouse}
  {\type & \sksp & \smouse }
\caption{Prediction using manual or produced training annotations. For each type, the mean of the maximum F1 scores for the proposed method (KSP) and the mouse-labeled case are shown.}\label{tab:learning}
\end{table}

\subsection{Experiment 3: Impact of coverage ratio and supervision}
When dealing with relatively large objects, it is quite possible that only the most salient parts of the object would be provided as locations, leaving large or homogeneous parts of the object unobserved. This is predominant in the case of the {\bf Tweezer} image sequence where a larger part of the shaft is often distant from any given $g_t$. In this experiment, we are interested in knowing to what degree the supervised 2D locations play a role in the quality of the segmented objects.

To estimate the impact of this effect, we evaluate the segmentation performance of our method as function of the coverage ratio (\ie ratio of the covered area over the total area of the object).
In this setting, we first selected a reference frame where the object appears in its entirety. We then manually select a set of positive superpixels from other frames so as to cover a pre-defined coverage ratio of the object surface. Note that only a single 2D location is still provided on each frame but that it is the entire set that specifies the coverage proportion of the object.
Each one of the $4$ sequences of the {{\bf Tweezer} dataset were assigned $5$ sets of 2D locations, each corresponding to approximately $\{20, 40, 60, 75, 90\}\%$ of the total area of the object (see Fig.~\ref{fig:coverage}, Left).

We report the F1 score with respect to the coverage ratio in Fig.~\ref{fig:coverage}(Right). As our method does not resort to frame-wise filling, we observe that the coverage ratio affects the F1-score as only ``seen'' regions end up being segmented. As such, attempting to recover the entire object over all frames from a few points remains extremely difficult. We note that even at $90\%$ coverage ratio, the F1 score is of roughly $0.82$ and not higher. As can be observed in the bottom right frame of Fig.~\ref{fig:coverage}(Left), this is largely due to oversegmentations introduced by the superpixels used.
\begin{figure}
\centering
\begin{subfigure}{.49\textwidth}
  \centering
  \includegraphics[width=\linewidth]{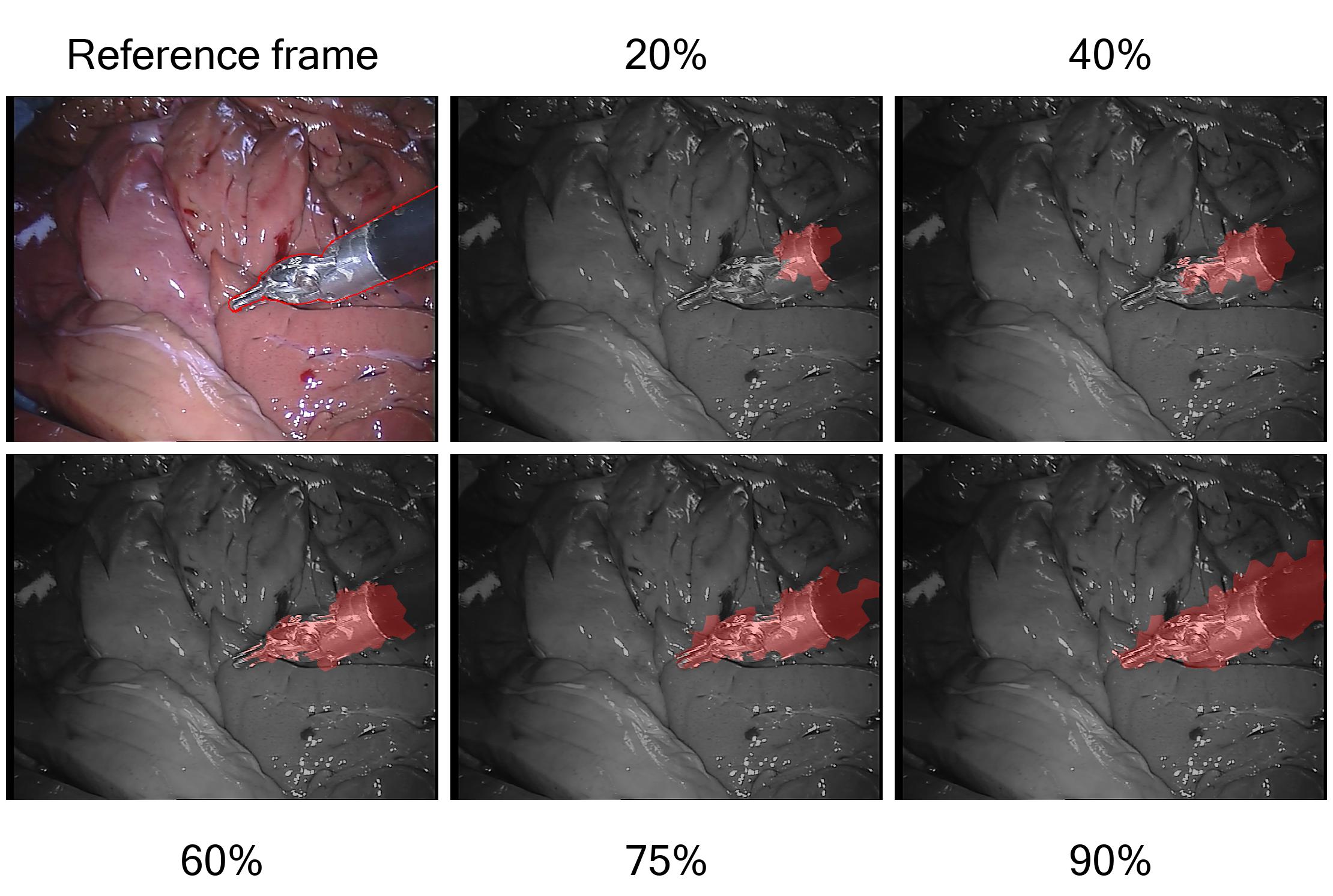}
  \label{fig:sub2}
\end{subfigure}
\begin{subfigure}{.49\textwidth}
  \centering
  \includegraphics[width=\linewidth]{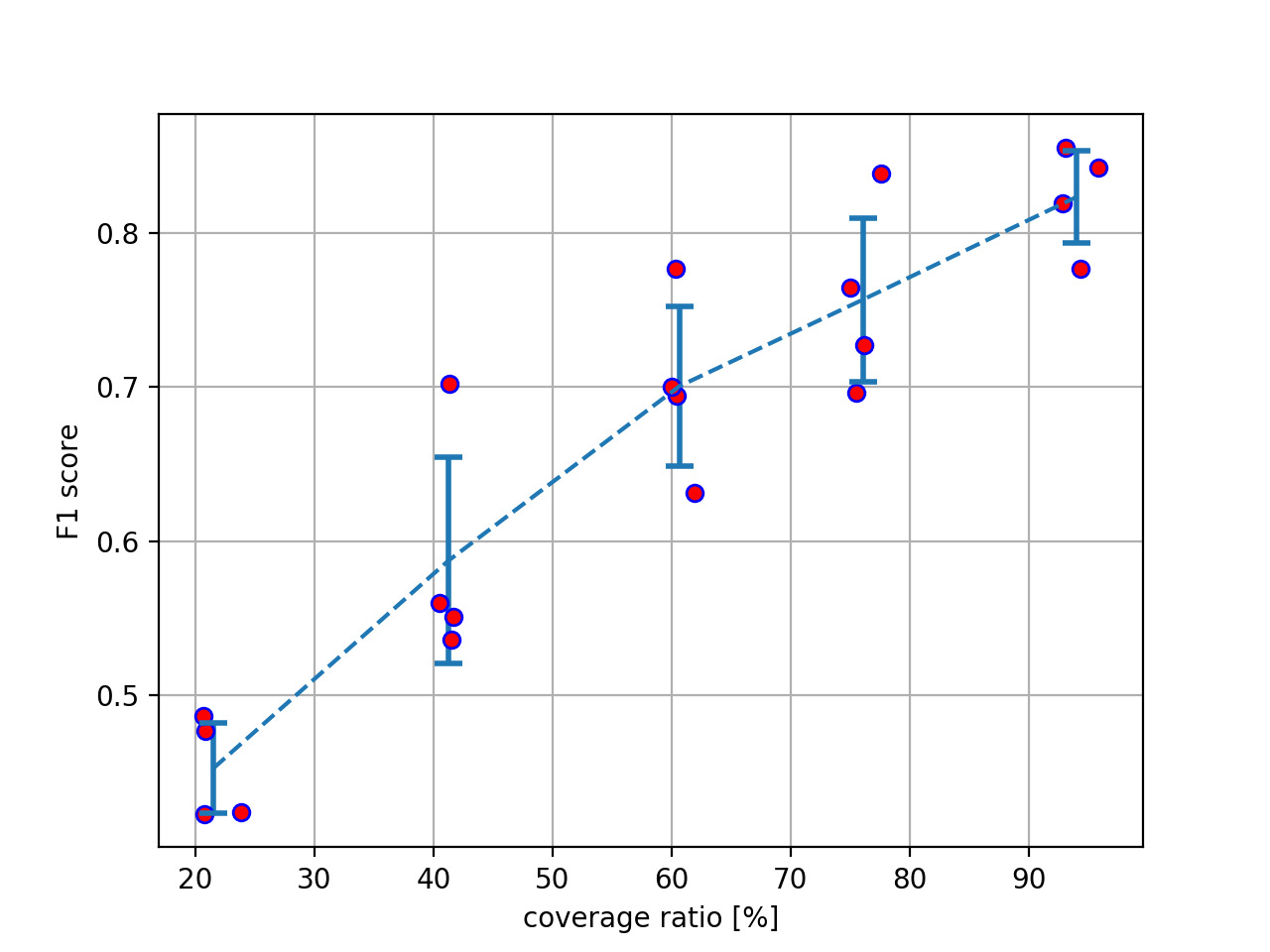}
  \label{fig:sub1}
\end{subfigure}%
\caption{(Left) Graphical examples of coverage ratio for a Tweezer sequence using each of the following ratios: $20$, $40$, $60$, $75$, and $90\%$. (Right) Boxplot of F1 scores with respect to coverage ratio on a Tweezer sequence. 
}
\label{fig:coverage}
\end{figure}

\begin{figure}[t]
\centering
\includegraphics[width=0.97\textwidth]{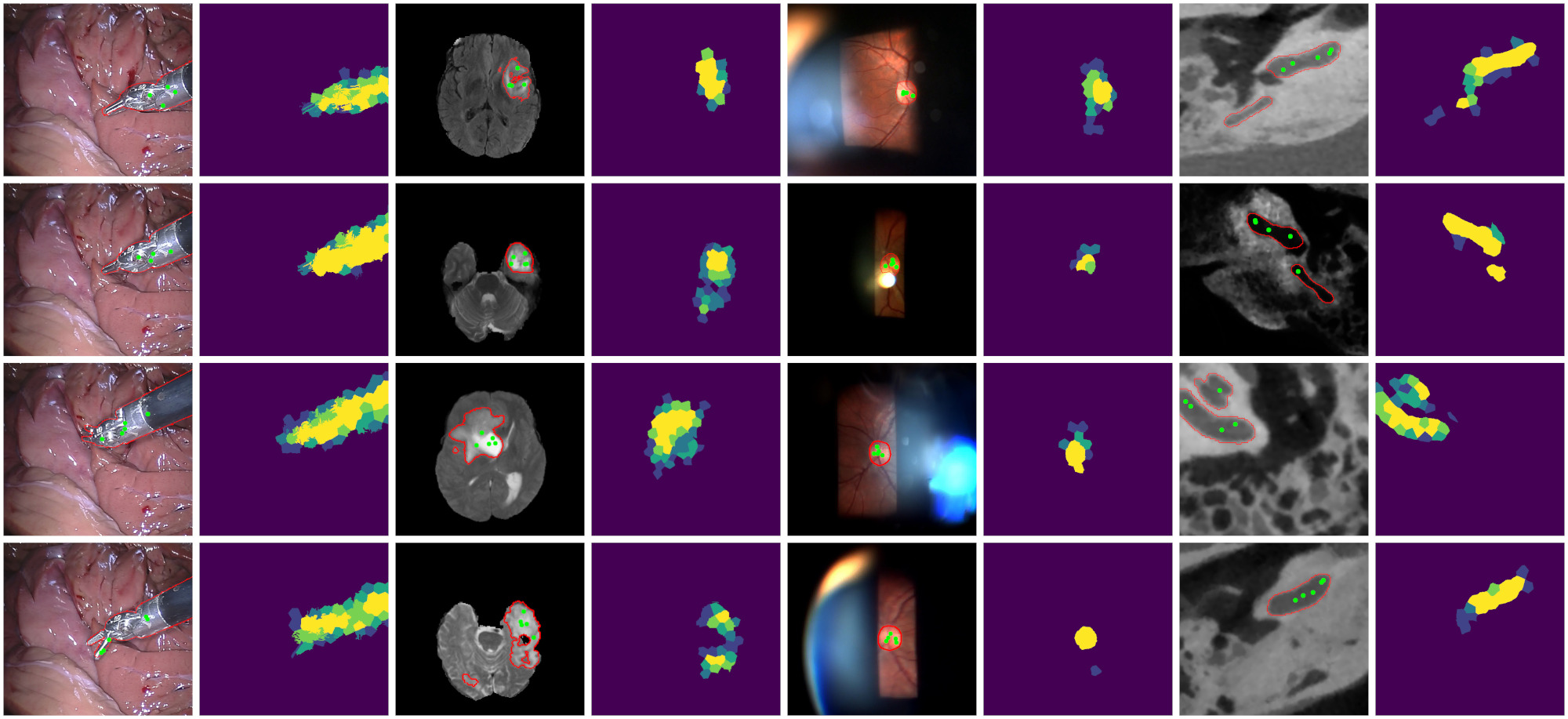}
\caption{Qualitative results of our approach when using different sets of supervised 2D locations. Columns 1, 3, 5, and 7 show the original image with highlighted ground truth contour in red. The 2D locations are in green. Columns 2, 4, 6, and 8 show the mean of all binary segmentations over $5$ sets of 2D locations.}
\label{fig:multigaze}
\end{figure}

In addition, Fig.~\ref{fig:multigaze} shows the mean segmentation over 5 different sets of provided 2D locations. We observe the largest inter-set variability on the {\bf Cochlea} dataset, where a standard deviation ranging from $5\%$ to $16\%$ is observed. The lowest variability is attained on the {\bf Tweezer} dataset within $2\%$ to $4\%$ range. The latter results can be explained by the fact that the {\bf Cochlea} dataset gives different possibilities as to which branch of the cochlea will be observed. In contrast, the {\bf Tweezer} dataset shows a stable appearance and shape throughout the sequences, thereby limiting the inter-user variability as each set covers roughly the same regions of the object.

\subsection{Experiment 4: Image-Object Features}
We also wish to assess the gain in performance of the proposed IOS method with respect to simpler approaches. In particular, we compare produced segmentation performances when using the following alternatives:
\begin{itemize}
\item[-]{\bf{U-Net:}} Using the same architecture as presented in Sec.~\ref{sec:features} in combination of a simple $L^2$ reconstruction loss,
    \begin{equation}
    \mathcal{L}' = \sum_{{I}_t \in \mathcal{I}} \sum_{k,l} \| I_t(k,l) - {\hat{I}}_t(k,l)\|^2.
    \label{eq:loss_features}
    \end{equation}
    \noindent
    This is similar to our object-image features, but without the object prior given by the 2D locations.
  \item[-]{\bf{OverFeat:}} We use the pre-trained CNN of~\cite{sermanet13}. The model is trained in a supervised classification setup on the ImageNet 2012 training set~\citep{deng09}, which contains about $1.2$ million natural images from $1000$ classes. In our setup, we use the provided \textit{fast} model, and extract square patches centered on the centroids $r_t^n$. We set the size of patches so as to include all pixels in superpixel $s_t^n$. The patch is then resized to ($231 \times 231$) and fed into the network to give a feature vector of size $4096$ at the output of the penultimate layer.
\end{itemize}

Table~\ref{tab:multigaze} shows the maximum F1-score and standard deviation when using different features in \KSP . Here we show the performance of each feature type over each sequence for each dataset. On average, IOS features provides superior performances over alternatives. In some cases (\eg Brain dataset), the standard U-Net or Overfeat features appear to perform better on two sequences in particular. Naturally, the performance variance with IOS features is higher given that they depend on the provided 2D locations. 

\begin{table}
\centering
\losscomp
\caption{Quantitative results of \KSPnb with different feature used on all datasets with five sets of 2D locations per sequence. Mean and standard deviationm F1 scores are given for 2D locations sets.}\label{tab:multigaze}
\end{table}

\subsection{Experiment 5: Impact of outliers and missing 2D locations}

Next, we look at the effect of the quality and quantity of provided 2D locations, that may vary in practice depending on various factors (\eg speed of object, gaze-tracker calibration accuracy etc.). To do this, we produce noisy versions of the 2D locations used in Sec.~\ref{sec:accuracy}. In particular, we generate for a given outlier proportion $\delta \in \{5, 10, 20, 40, 50 \} \%$, three sets of 2D locations. The first samples outliers uniformly on the background, the second samples uniformly on a neighborhood of the object at a distance of $5\%$ (normalized w.r.t. largest dimension of the image), and the last samples at a distance of $10\%$.
Examples for the last two cases are illustrated in Fig.~\ref{fig:neigh_5} and~\ref{fig:neigh_10}, respectively.
We also show the case where a proportion of 2D locations are missing entirely.
Fig.~\ref{fig:outliers} depicts the F1 score with respect to the proportion of outlier or missing locations on the {\bf Tweezer} sequence \#1.
We report for missing 2D locations a maximum decrease in F1 score of $12\%$ when $40\%$ of locations are missing. This showcases the robustness of our method brought by the global data association optimization. Outliers however impact our method more severely as for incorrect 2D locations of up to $ 5\%$ decrease performances by $14\%$ with $20\%$ of incorrect locations. Similarly, with errors at a maximum of the $10\%$ distance, the performance decreases by $18\%$ when $40\%$ of the locations are corrupted. For severe corruptions of $50\%$ over the entire background, then the performance drops by $45\%$. 
In general however, we note that the performance remains acceptable up to a $40\%$ outlier proportion. This is explained by the fact that our foreground model effectively penalizes such outliers through its bagging component.

\begin{figure}
\centering
\begin{subfigure}[t]{.25\textwidth}
  \centering
  \includegraphics[width=\linewidth]{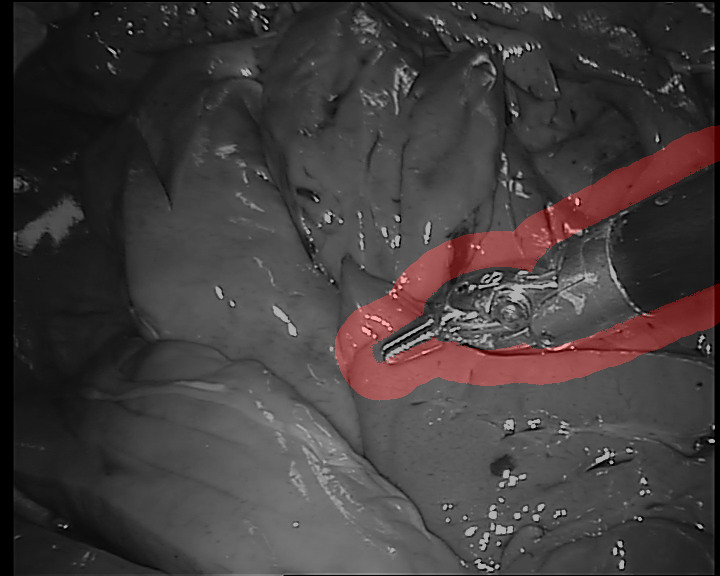}
  \caption{\label{fig:neigh_5}}
\end{subfigure}
\begin{subfigure}[t]{.25\textwidth}
  \centering
  \includegraphics[width=\linewidth]{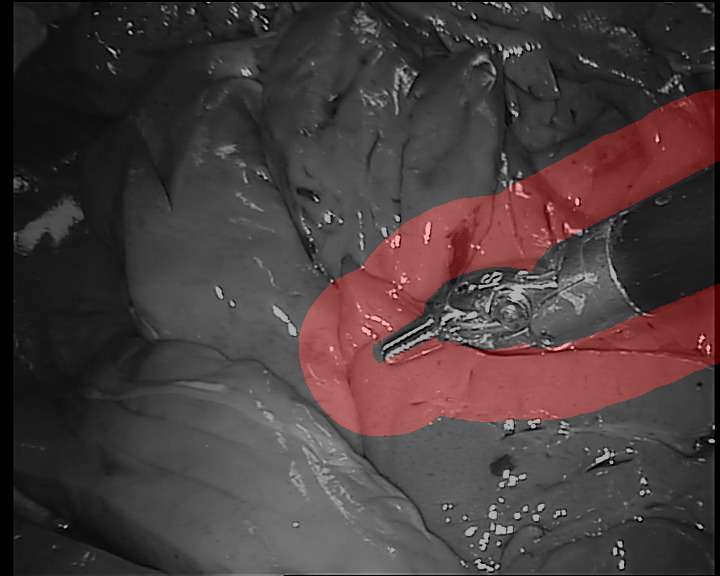}
  \caption{\label{fig:neigh_10}}
\end{subfigure}%
\begin{subfigure}[t]{.4\textwidth}
  \centering
  \includegraphics[width=\linewidth]{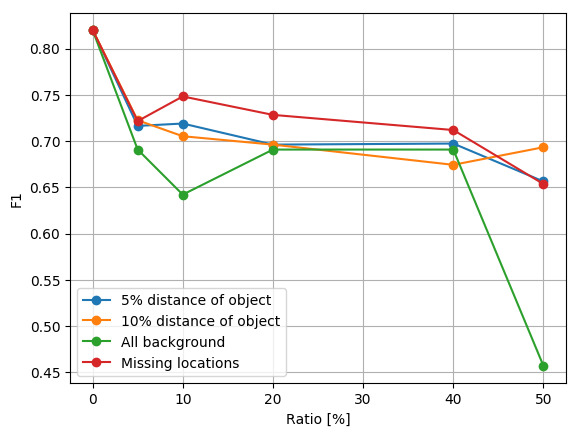}
  \caption{\label{fig:outliers}}
\end{subfigure}%
\caption{
(a and b) Example frames with outlier regions highlighted in red corresponding to distance of $5\%$ and $10\%$, respectively. (c) F1 scores with respect to the corrupted proportion of provided 2D locations.\label{fig:outliers_missing}
}
\end{figure}
\section{Conclusion}
\label{sec:conclusion}
In this paper, we presented a framework that allows for pixel-wise segmentation of objects of interest to be generated from sparse sets of 2D locations in video and volumetric image data. In this context, we have provided a strategy that produces an object segmentation by formulating the task as a global multiple-object tracking problem and solving it using an efficient K-shortest paths algorithm. Using an object model estimated from the sparse set of locations, we iteratively refine our solution by progressively improving our object model. To do this effectively, we introduce the use of image-object specific (IOS) features for our purpose and which are generated from an autoencoder that leverages the 2D object locations as a soft prior. 

We show in our experiments that our approach is capable of reliably segmenting complex objects of interest over a wide range of image sequences and 3D volumes. Unlike previous methods, our \KSP ~method does not assume that the object is of a given size or any information about the background is known. Yet by combining our multi-path tracker and our object model, we achieve superior segmentation results compared to a number of state-of-the-art methods. Beyond this, we show that our results are stable under a number of conditions including the specific nature of the provided 2D locations, even when these are collected at framerate using a low cost gaze tracker.

While we demonstrate in our experiments that generated segmentations could be used to train supervised machine learning segmentation methods without suffering too greatly, we show that the performance of our method does depend on the spatial coverage of the provided 2D locations. In the future, we will look to further exploit inter-frame consistency to refine segmentations, in particular to recover fine object details. In addition, so far we have assumed that only a single object is within the data volume. As such, we will look to overcome this limitation and investigate how transfer learning strategies could benefit both feature extraction and segmentation towards this end. Last, in our current set-up, the size of the superpixels used can negatively impact the segmentation produced. To limit the impact of this shortcoming, the use of strategies that refine or merge superpixels to produce more accurate final results will be investigated.

\section*{Acknowledgements}
This work was supported in part by the Swiss National Science Foundation Grant 200021 162347 and the University of Bern.


\appendix
\section{Edge-disjoint K-shortest paths}
\label{sec:ksp}

For the sake of completeness, we provide a summary of the edge-disjoint K-shortest paths algorithm implemented in this work while the full version is given in~\cite{suurballe74}.
Alg.~\ref{alg:ksp} describes the pseudo code of what follows.
Given a directed acyclic graph $G$, the edge-disjoint K-shortest paths algorithm iteratively augments the set of $l$ shortest paths $P_{l}$, to obtain the optimal set $P_{l+1}$. Starting with $l=0$, we use a generic shortest-path algorithm to compute $P_0=\{p_0\}$.
In practice, we use the Bellman-Ford's algorithm~\citep{bellman58}, which is adequate to cases where edges have negative costs. We then perform two kinds of transformations on $G$.

\begin{itemize}
\item[-]  {\bf{Reverse operation:}} The direction and algebraic signs of edge costs occupied by path(s) of $P_{l}$ are reversed.
\item[-] {\bf{Edge costs transform:}}
The generic shortest-paths algorithm gives for every node the cost of the shortest path from the source. We perform a cost transformation step to make all edges of our graph non-negative. This allows us to then use the more efficient Dijkstra's single source shortest path algorithm~\citep{dijkstra59}, which requires non-negatives edge costs as well.
In particular, we let $v_t^n$ and $w_t^n$ be the input and output nodes of tracklets $\mathcal{T}_t^n$, and let $L(v_t^n)$ be the cost of the shortest path from the source to node $v_t^n$. We apply $\forall m,n,t$:
\begin{subequations}
\label{eq:cost_transform}
\begin{align}
&C_t^{n} \coloneqq C_t^n + L(v_t^n) - L(w_t^n)\label{eq:cost_transform_tracklet}\\
&C_{t-1}^{m,n} \coloneqq C_{t-1}^{m,n} + L(w_{t-1}^m) - L(v_{t}^n)\label{eq:cost_transform_transition}\\
&C_t^{\mathcal{E}_t,n} \coloneqq 0 \label{eq:cost_transform_entrance}\\
&C_t^{n,\mathcal{X}} \coloneqq C_t^{n,\mathcal{X}} + L(w_t^n) - L(\mathcal{X}). \label{eq:cost_transform_sink}
\end{align}
\end{subequations}
\end{itemize}

On this modified graph, we compute the interlacing path $\tilde{p}_0$. The set $P_{1}$ is obtained by {\it augmenting} $P_0$ with $\tilde{p}_0$. Concretely, we assign a negative label to the edges of $\tilde{p}_0$ that are directed towards the source, and a positive label otherwise. We then construct the optimal pair of paths $P_{1}$ by adding positive edges of $\tilde{p}$ to $P_0$ and removing negative edges from $P_0$, as shown on Fig.~\ref{fig:augment}. The next iterations follow the same procedure. Note that in general, $\tilde{p}_l$ can interlace several paths of $P_l$.

Hence, our algorithm runs Dijkstra's single source shortest path $K$ times. We therefore have a complexity time linear with $K$, \ie in a worst case scenario: $\mathcal{O} \left( K(E + V \cdot \log \, V) \right)$, with $K, V, E$ the number of path sets, nodes, and edges, respectively.

\begin{figure}[t]
\centering
\includegraphics[width=0.49\textwidth]{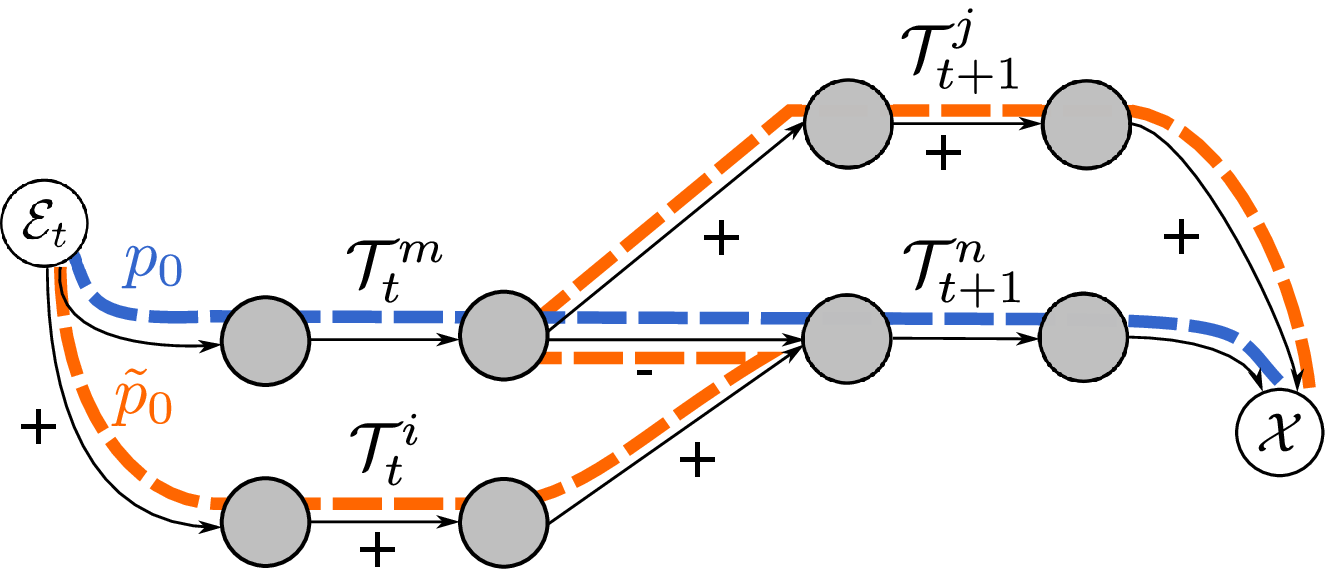}
\includegraphics[width=0.49\textwidth]{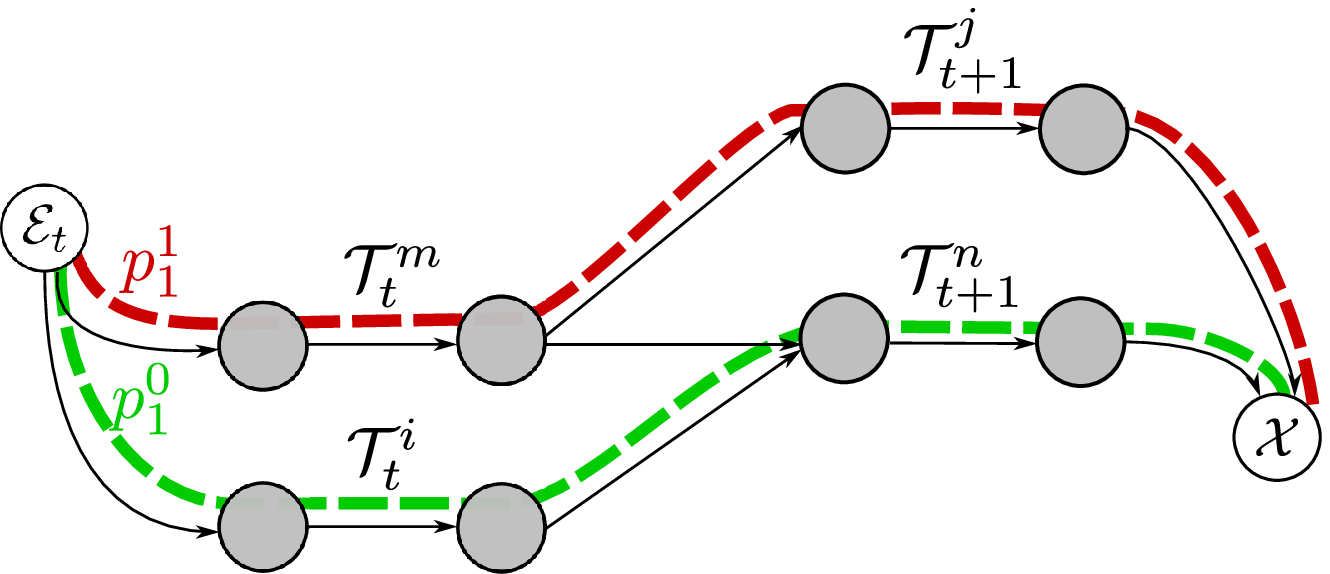}
\caption{Illustration of the interlacing and augmentation procedure for $K$=2. (Left) $p_0$ is the (single) shortest-path of set $P_0$. $\tilde{p}_0$ is the shortest interlacing path obtained after inverting the direction and algebraic sign of edge costs of $P_0$. Positive and negative labels are assigned to the edges of $P_0$. (Right) The optimal set $P_1=\{ p_1^0,p_1^1 \}$ is obtained by removing the edges with negative labels from $p_0$, and adding positive labels.}\label{fig:augment}
\end{figure}

\SetArgSty{textnormal}
\begin{algorithm}[t]
\SetKwInOut{Input}{Input}
\SetKwInOut{Output}{Output}
\DontPrintSemicolon
\Input{$G$: Directed Acyclic Graph constructed as in Sec.~\ref{sec:solving}}
\Output{$P$: Set of K-shortest paths}
$p_0 \gets$ \texttt{bellman\_ford\_shortest\_paths}$(G)$\;
$P_0 \gets\{p_0\} $\;
\For{$l\gets0$ \KwTo $l_{max}$ }{
 \If{$l \neq 0$}{
   \If{$\text{cost}(P_l) \geq \text{cost}(P_{l-1})$}{
     \Return $P_{l-1}$
     }
 }
   $G_r \gets$ \texttt{reverse}$(G,P_l)$\tcp*{Reverse edges directions and algebraic signs}
   $G_r^+ \gets$ \texttt{edge\_costs\_transform}$(G_r)$\tcp*{As in Eq.~\eqref{eq:cost_transform}}
   $\tilde{p}_l \gets$ \texttt{dijkstra\_shortest\_paths}$(G_r^+)$\tcp*{Returns interlacing path}
   $P_{l+1} \gets$ \texttt{augment}$(P_l, \tilde{p}_l)$\tcp*{As on Fig.~\ref{fig:augment}}

   }
\caption{K-shortest paths algorithm.\label{alg:ksp}}
\end{algorithm}


\section*{References}
{\small
\bibliography{refs}
}

\end{document}